\documentclass[8pt, a4paper, twocolumn]{extarticle}

\usepackage{amssymb}
\usepackage{authblk}
\usepackage{amsmath}
\usepackage{lipsum}
\usepackage{bm}
\usepackage{graphicx}
\usepackage{algorithm} 
\usepackage{algorithmic}  
\usepackage[algoruled,algo2e]{algorithm2e}
\usepackage{setspace}
\usepackage{enumerate}
\usepackage{sidecap}
\usepackage{wrapfig}
\usepackage{import}
\usepackage{float}
\usepackage{subcaption}

\usepackage{hyperref}
\hypersetup{colorlinks=true, urlcolor=blue, linkcolor=blue, citecolor=blue}
\usepackage[margin=0.75in]{geometry}
\usepackage[dvipsnames]{xcolor}

\usepackage{mathtools}
\usepackage{scalerel}
\newcommand{\rfpe}[2] {
  \varphi^{#2} \mathrlap{}_{\!\! \scaleto{#1}{5.9pt}}}

\usepackage{tikz}
\usepackage{pgfplots}
\usepgfplotslibrary{statistics}
\usetikzlibrary{matrix}
\usepgfplotslibrary{patchplots}
\usetikzlibrary{calc,3d}
\definecolor{customblueinksc1}{RGB}{28, 28, 175} 

\usepackage[utf8]{inputenc}
\DeclareUnicodeCharacter{2212}{−}
\usepgfplotslibrary{groupplots,dateplot, units}
\usetikzlibrary{patterns,shapes.arrows}
\pgfplotsset{compat=newest}
\newlength\figH
\newlength\figW
\definecolor{color0}{HTML}{485899}
\definecolor{color1}{HTML}{c25955}
\definecolor{color2}{HTML}{6c736c}

\renewenvironment{thebibliography}[1]{
  \begin{oldthebibliography}{#1}
    \setlength{\itemsep}{0em}
    \setlength{\parskip}{0em}
}
{
  \end{oldthebibliography}
}

\newcommand{\tmp}{\textcolor{black}}

\newcommand{\vect}[1]{\bm{#1}}
\newcommand{\mat}[1]{\bm{#1}}

\newcommand{\diff}{\mathop{}\!\mathrm{d}}

\newcommand{\hp}{\vect{\kappa}}
\newcommand{\rfp}{\varphi}

\newcommand{\bx}{\bs{x}}

\newcommand{\bt}{\bs{\theta}}

\newcommand{\be}{\begin{equation}}
\newcommand{\ee}{\end{equation}}
\newcommand{\bc}{\begin{center}}
\newcommand{\ec}{\end{center}}
\newcommand{\bd}{\begin{description}}
\newcommand{\ed}{\end{description}}
\newcommand{\bi}{\begin{itemize}}
\newcommand{\ei}{\end{itemize}}

\newcommand{\bs}{\boldsymbol}

\def\RR{ \mathbb R}
\newcommand{\refeqp}[1]{Eq. (\ref{#1})}

\newcommand{\bmat}{\begin{pmatrix}}
\newcommand{\emat}{\end{pmatrix}}
\newcommand{\bsmat}{\left(\begin{smallmatrix}}
\newcommand{\esmat}{\end{smallmatrix}\right)}
\newcommand{\bes}{\begin{equation}\begin{split}}
\newcommand{\ees}{\end{split}\end{equation}}

\author[1]{Maximilian Rixner}
\author[*,1,2]{Phaedon-Stelios Koutsourelakis}
\date{}   
\affil[1]{Technical University of Munich, Faculty of Mechanical Engineering, Professorship of Continuum Mechanics, www.contmech.mw.tum.de }
\affil[2]{Munich Data Science Institute (MDSI - Core Member), www.mdsi.tum.de }
\title{Self-supervised optimization of {\em random} material microstructures in the small-data regime}

\newcommand{\abstractText}{\noindent
While the forward and backward modeling of the  process-structure-property  chain has received a lot of attention from the materials community, fewer efforts have taken into consideration uncertainties. Those arise from a multitude of sources and their quantification and integration in the inversion process are essential in meeting the materials design objectives.
The first contribution of this paper is a flexible, fully probabilistic formulation of such optimization problems that accounts  for the uncertainty in the process-structure and structure-property linkages and enables the identification  of optimal, high-dimensional, process parameters.  We employ a probabilistic, data-driven surrogate for the structure-property link which expedites computations and enables handling of non-differential objectives. We couple this with a novel  active learning strategy, i.e. a self-supervised collection of data, which significantly improves accuracy while requiring small amounts of training data. We demonstrate its efficacy in optimizing the mechanical and thermal properties of two-phase, random  media but envision its applicability encompasses a wide variety of microstructure-sensitive design problems.}

\begin{document}


\twocolumn[
  \begin{@twocolumnfalse}
    \maketitle
    \begin{abstract}
      \abstractText
      \newline
      \newline
    \end{abstract}
  \end{@twocolumnfalse}
]


\section*{Introduction}  

Inverting the process-structure-property  (PSP) relationships represents a grand challenge in materials science as it holds the potential of expediting the design of new materials with superior performance \cite{mgi_2011,mcdowell_integrated_2009}. 

While significant progress has been  made in the forward and backward modeling of the process-structure and structure-property linkages and  in capturing the nonlinear and multiscale processes involved \cite{arroyave_systems_2019}, much fewer efforts have attempted to integrate uncertainties which are an indispensable component of materials' analysis and design \cite{chernatynskiy_uncertainty_2013,honarmandi_uncertainty_2020} since a) process variables do not fully determine the resulting microstructure but rather a probability distribution on microstructures  \cite{liu_xuan_nasa_2018}, b) noise and incompleteness are characteristic of experimental data that are used to capture  process-structure (most often) and  structure-property relations \cite{bock_review_2019}, c) models employed for the process-structure or structure-property links are often stochastic and there is uncertainty in their parameters or form, especially in multiscale formulations \cite{panchal_key_2013}, and d) model compression and dimension reduction employed in order to gain efficiency unavoidably leads to some loss of information which in turn gives rise to predictive uncertainty \cite{grigo_bayesian_2019}.
As a result, microstructure-sensitive properties can exhibit stochastic variability which should be incorporated in the design objectives.

(Back-)propagating uncertainty through complex and potentially multiscale models poses significant computational difficulties \cite{zabaras_scalable_2008}.
Data-based surrogates can alleviate these as long as the number of training data, i.e. the number of solves of the complex  models they would substitute, is kept small. 
In this small-data setting additional  uncertainty arises due to the predictive inaccuracy of the surrogate. Quantifying it can not only lead to more accurate estimates but also guide the acquisition of additional experimental/simulation data.

We note that problem formulations based on Bayesian Optimization  \cite{frazier_bayesian_2016,zhang_bayesian_2020,jung_microstructure_2020}) account for  uncertainty in the objective solely  due to the imprecision of the surrogate and not due to the aleatoric, stochastic variability of the underlying microstructure.
In the context of optimization/design problems in particular,  a globally-accurate surrogate  would be redundant. It would suffice to have a surrogate that can reliably drive the optimization process to the vicinity of the local optimum (or optima) and can  sufficiently resolve this (those) in order to identify the optimal value(s).
Since the location of optima is unknown a priori this necessitates adaptive strategies where the surrogate-training and optimization are intertwined. 

\begin{figure*}[t!]
    \centering
    \includegraphics[scale=0.99]{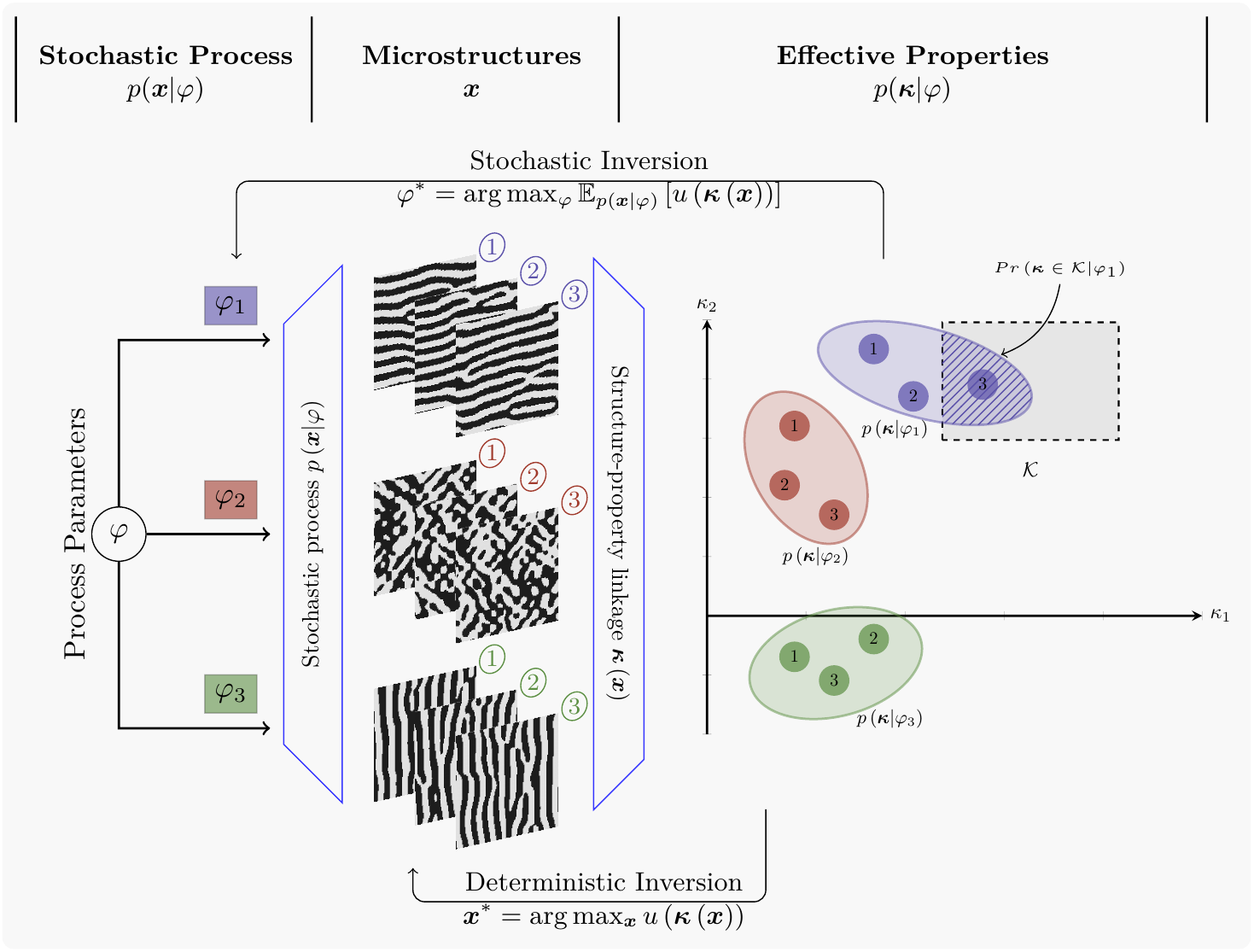}
    \caption{ \textbf{Conceptual overview.} 
    Given stochastic process-structure and structure-property links, we identify the  process parameters $\varphi^*$ which maximize the expected utility $\mathbb{E}_{p \left( \bm{x} \middle| \varphi \right)} \left[ u \left( \hp \right) \right]$ (Illustration based on the specific case  $u \left( \hp \right) = \mathbb{I}_{\mathcal{K}} \left( \hp \right) $). (Micro)Structures $\bx$ arise from a stochastic process through the density $p \left( \bm{x} \middle| \varphi \right)$ that depends on the process parameters $\rfp$. A data-driven surrogate is employed to predict properties $\hp$ which introduces additional uncertainty.
    }
    \label{fig:overview}
\end{figure*}

We emphasize that unlike successful efforts e.g. in topology optimization \cite{chen_machine_2019} or general heterogeneous media \cite{torquato_optimal_2010} which find a single, optimal microstructure that maximizes some property-based objective, our goal is more ambitious but also more consistent with the physical reality. We attempt to find the optimal distribution of microstructures from the ones that are realizable from a set of processing conditions (Fig. \ref{fig:overview}). To address the computational problem arising from the presence of uncertainties, we recast the stochastic optimization as a probabilistic inference task and employ approximate inference techniques based on Stochastic Variational Inference (SVI, \cite{hoffman_stochastic_2013}).

In terms of the stochastic formulation of the problem, our work most closely resembles that of \cite{tran_solving_2021} where they seek to identify a  probability density on microstructural features  which would yield a target probability density on the corresponding properties. While this poses a challenging optimization problem, producing a probability density on microstructural features
 does not provide unambiguous design guidelines.
In contrast, we operate on (and average over) the whole microstructure and consider a much wider range of design objectives.
In \cite{nosouhi_dehnavi_framework_2020} random microstructures were employed but their macroscopic properties were insensitive to their random variability (due to scale-separation) and low-dimensional parametrizations of the two-point correlation function were optimized using gradient-free tools.
In a similar fashion, in \cite{acar_stochastic_2017,acar_stochastic_2019} analytic, linear models are employed which given small and Gaussian uncertainties on the macroscopic properties, find the underlying orientation distribution function (ODF) of the crystalline microstructure. 
In \cite{liu_predictive_2015,paul_microstructure_2019}, averaged macroscopic properties (ignoring the effects of crystal size and shape) are computed with respect to the ODF of the polycrystalline microstructure and  on the basis of their targeted values, the corresponding ODF is found.  While data-based surrogates were also employed, the problem formulation did not attempt to quantify the effect of microstructural uncertainties.

In terms of surrogate development, in this work we focus on the microstructure-property link and  consider random, binary microstructures, the distribution of which depends on some processing-related parameters. We develop active learning strategies that are tailored to the optimization objectives. The latter account for the stochasticity of the material properties (as well as the predictive uncertainty of the surrogate) i.e. we enable the solution of   optimization under uncertainty  problems.
The methodological framework presented  enables the control of  a \textit{stochastic process} giving rise to the discovery  of  {\em random} heterogeneous materials,  that  exhibit favourable properties according to some notion of optimality.

\section*{Methods}

\begin{figure*}[t]\centering 
	\includegraphics[width=0.70 \linewidth]{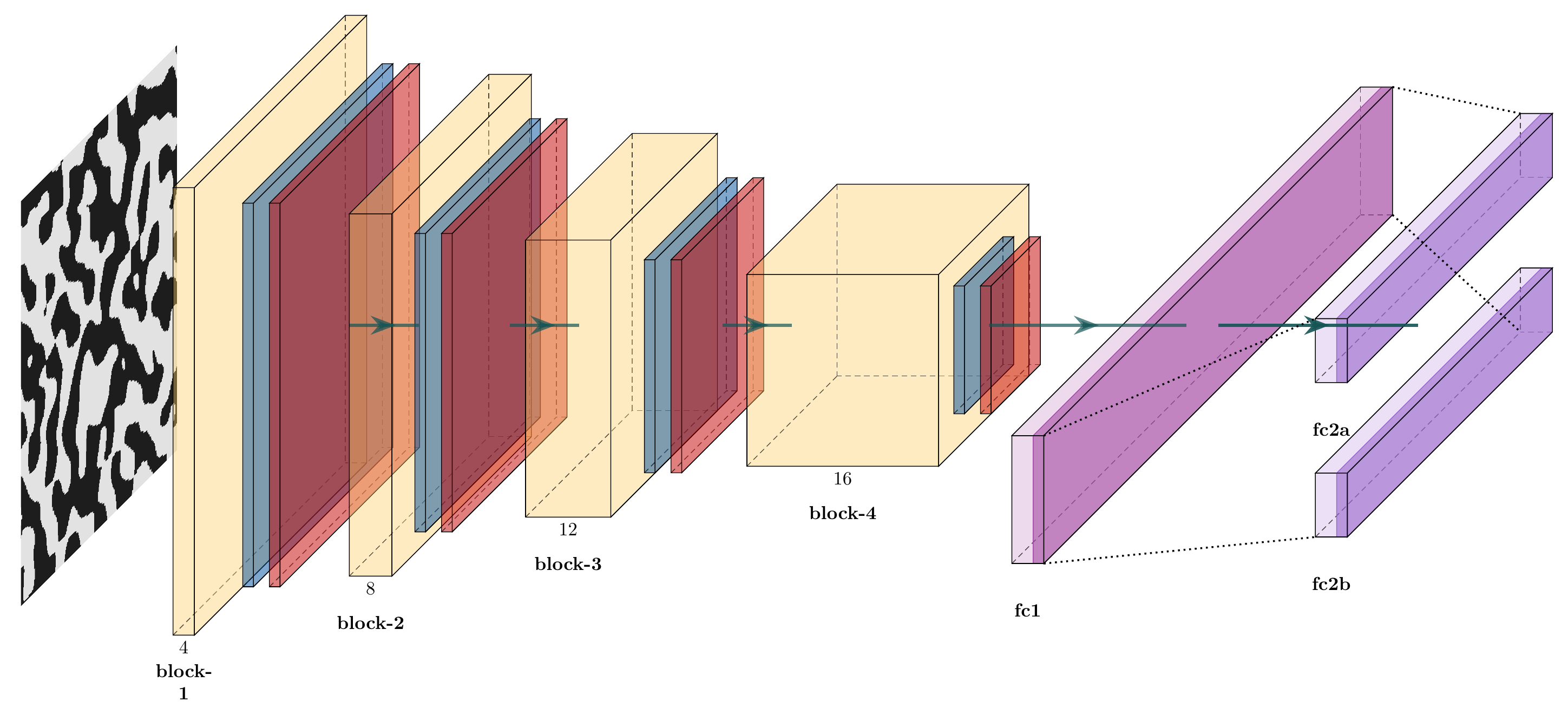}
	\caption{\textbf{Architecture of the convolutional-neural-network surrogate for property $\hp$ prediction}. Features are extracted from the microstructure $\bm{x}$ using a sequence of $4$ blocks (each comprised of a sequence of convolutional layer, non-linear activation function and pooling), where in each block the size of the feature map is reduced, while the depth of the feature map increases. Fully connected feedforward layers map the extracted convolutional features to the mean $\bm{m}_{\bm{\theta}} \left( \bm{x} \right)$ and the covariance $\mat{S}_{\bm{\theta}} \left( \bm{x} \right)$ of the predictive  Gaussian distribution $p_{\mathcal{M}} \left( \hp \middle| \bm{x}, \bm{\theta} \right) = \mathcal{N} \left( \hp \middle|~ \bm{m}_{\bm{\theta}} \left( \bm{x} \right), \mat{S}_{\bm{\theta}} \left( \bm{x} \right) \right)$, where $\bm{\theta}$ denotes the neural network parameters.}
	\label{fig:CNN_architecture}
\end{figure*}

A conceptual overview of the proposed stochastic-inversion framework is provided in Fig. \ref{fig:overview} and is contrasted with deterministic formulations. 
We present the main building blocks and modeling assumptions and  subsequently define the  optimization problems of interest. We then discuss associated  challenges, algorithmic steps and   conclude this section with  details regarding the probabilistic surrogate model and the active learning strategy.

We define the following variables/parameters:
\bi
\item process parameters $\varphi \in \RR^{d_{\phi}}$: These are the optimization variables and can parametrize actual processing variables (e.g. chemical composition, annealing temperature) or statistical descriptors (e.g. ODF) that might be linked to the processing.  In general, high-dimensional $\varphi$ would need to be considered which  can afford greater flexibility in the design process.

\item {\em random} microstructures $\bx$: This is in general a very high-dimensional vector that represents the microstructure with the requisite detail  to predict its properties. In the numerical illustrations which involve two-phase media in $d=2$ dimensions represented on a uniform grid with $N_p$ subdivisions per dimension,  $\bx \in \{0,1\}^{N_p^d}$
 consists of binary variables which indicate the material phase of each of the  pixels (Figure \ref{fig:multiphysics_microstructures}). We  emphasize that $\bx$ is a {\em random vector} due to the stochastic variability of microstructures even in cases where $\varphi$ is the same (see process-structure link below).

\item properties $\vect{\kappa}$: This vector represents the material properties of interest which depend on the microstructure $\bx$. We denote this dependence with some abuse of notation as $ \vect{\kappa}(\bx)$ and discuss it in the structure-property link below. Due to this dependence,  $\vect{\kappa} \in \RR^{d_k}$ will also be a random vector. In the numerical illustrations   $\vect{\kappa}$ consists of mechanical and thermal, effective (apparent) properties.
\ei
Furthermore, we note:
\bi
\item process-structure link: We denote the dependence between $\varphi$ and $\bx$  with the conditional density $p(\bx|\varphi)$ (Figure \ref{fig:overview}) that reflects the fact that processing parameters do not in general uniquely determine the microstructural details. 
Formally  experimental data \cite{popova_process-structure_2017}  and/or models  \cite{lee_fast_2021}\footnote{The binary microstructures considered for our numerical illustrations could arise from the solution of the Cahn-Hilliard equation describing phase separation occurring in a binary alloy under thermal annealing}  would need to be  used to determine $p(\bx|\varphi)$. We also note that no a-priori dimensionality reduction is implied, i.e., the full microstructural details are retained and employed in the property-predicting, high-fidelity models. In this work, we assume the process-structure link $p \left( \bm{x} \middle| \varphi \right)$ is given, and its particular form for the binary media examined is explained in the sequel.   

\item structure-property link: The calculation of the properties $\hp$ for a given microstructure $\bx$ involves in general the solution of a stochastic or deterministic, complex, high-fidelity model (in our numerical illustrations, this consists of  partial differential equations (PDEs)). We denote the corresponding conditional density as $p(\hp | \bx)$, which in the case of a deterministic model degenerates to a Dirac-delta. For optimization purposes, repeated such computations are needed  and especially in high-dimensional settings  derivatives of the properties with respect to $\bx$  are  also required in order to drive the search. Such derivatives  might be either unavailable (e.g. when $\bx$ is binary as above),  or, at the very least, will add to the computational burden.  
To overcome this major efficiency hurdle we advocate the use of a data-driven surrogate model. We denote with $\mathcal{D}$ the training data (i.e. pairs of inputs-microstructures and outputs-properties $\hp(\bx)$) and explain in the sequel how these are selected (see section on Active Learning). We employ a {\em probabilistic} surrogate model denoted by $\mathcal{M}$  and use  $p_{\mathcal{M}}(\hp| \bx, \mathcal{D})$ to indicate its predictive density (we will elaborate on the reasons for using a  {\em probabilistic} surrogate in the subsequent sections).
\ei

\begin{figure*}[t!]
\centering
\begin{minipage}{0.3\textwidth}
  \centering
\resizebox{0.90\linewidth}{!}{
\begingroup%
  \makeatletter%
  \providecommand\color[2][]{%
    \errmessage{(Inkscape) Color is used for the text in Inkscape, but the package 'color.sty' is not loaded}%
    \renewcommand\color[2][]{}%
  }%
  \providecommand\transparent[1]{%
    \errmessage{(Inkscape) Transparency is used (non-zero) for the text in Inkscape, but the package 'transparent.sty' is not loaded}%
    \renewcommand\transparent[1]{}%
  }%
  \providecommand\rotatebox[2]{#2}%
  \newcommand*\fsize{\dimexpr\f@size pt\relax}%
  \newcommand*\lineheight[1]{\fontsize{\fsize}{#1\fsize}\selectfont}%
  \ifx\svgwidth\undefined%
    \setlength{\unitlength}{174.62322831bp}%
    \ifx\svgscale\undefined%
      \relax%
    \else%
      \setlength{\unitlength}{\unitlength * \real{\svgscale}}%
    \fi%
  \else%
    \setlength{\unitlength}{\svgwidth}%
  \fi%
  \global\let\svgwidth\undefined%
  \global\let\svgscale\undefined%
  \makeatother%
  \begin{picture}(1,0.69330564)%
    \lineheight{1}%
    \setlength\tabcolsep{0pt}%
    \put(0,0){\includegraphics[width=\unitlength,page=1]{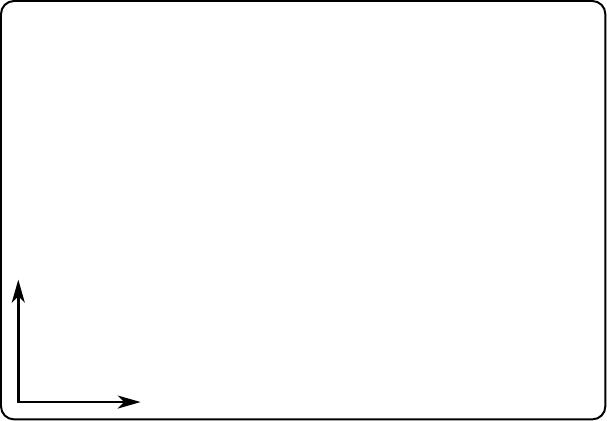}}%
    \put(0.24850682,0.02319553){\color[rgb]{0,0,0}\makebox(0,0)[lt]{\lineheight{1.25}\smash{\begin{tabular}[t]{l}$\kappa_1$\end{tabular}}}}%
    \put(0.01995366,0.25251568){\color[rgb]{0,0,0}\makebox(0,0)[lt]{\lineheight{1.25}\smash{\begin{tabular}[t]{l}$\kappa_2$\end{tabular}}}}%
    \put(0,0){\includegraphics[width=\unitlength,page=2]{objectivecandidates_a.pdf}}%
    \put(0.63485329,0.57297165){\color[rgb]{0,0,0}\makebox(0,0)[lt]{\lineheight{1.25}\smash{\begin{tabular}[t]{l}$p(\bm{\kappa} | \varphi^*)$\end{tabular}}}}%
    \put(0.53622736,0.1956249){\color[rgb]{0,0,0}\makebox(0,0)[lt]{\lineheight{1.25}\smash{\begin{tabular}[t]{l}\textcolor{customblueinksc1}{$\mathcal{K}$}\end{tabular}}}}%
    \put(0,0){\includegraphics[width=\unitlength,page=3]{objectivecandidates_a.pdf}}%
  \end{picture}%
\endgroup%
}
\subcaption{Target domain $\mathcal{K}$}\label{fig:objectivetypes-a}
\end{minipage}%
\begin{minipage}{0.3\textwidth}
  \centering
\resizebox{0.90\linewidth}{!}{
\begingroup%
  \makeatletter%
  \providecommand\color[2][]{%
    \errmessage{(Inkscape) Color is used for the text in Inkscape, but the package 'color.sty' is not loaded}%
    \renewcommand\color[2][]{}%
  }%
  \providecommand\transparent[1]{%
    \errmessage{(Inkscape) Transparency is used (non-zero) for the text in Inkscape, but the package 'transparent.sty' is not loaded}%
    \renewcommand\transparent[1]{}%
  }%
  \providecommand\rotatebox[2]{#2}%
  \newcommand*\fsize{\dimexpr\f@size pt\relax}%
  \newcommand*\lineheight[1]{\fontsize{\fsize}{#1\fsize}\selectfont}%
  \ifx\svgwidth\undefined%
    \setlength{\unitlength}{174.62322831bp}%
    \ifx\svgscale\undefined%
      \relax%
    \else%
      \setlength{\unitlength}{\unitlength * \real{\svgscale}}%
    \fi%
  \else%
    \setlength{\unitlength}{\svgwidth}%
  \fi%
  \global\let\svgwidth\undefined%
  \global\let\svgscale\undefined%
  \makeatother%
  \begin{picture}(1,0.69330564)%
    \lineheight{1}%
    \setlength\tabcolsep{0pt}%
    \put(0,0){\includegraphics[width=\unitlength,page=1]{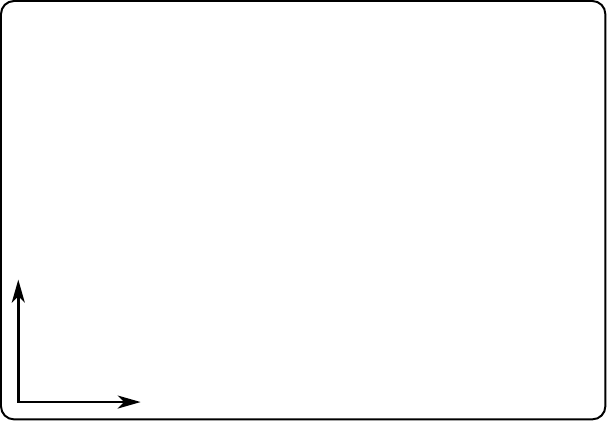}}%
    \put(0.24850682,0.02319553){\color[rgb]{0,0,0}\makebox(0,0)[lt]{\lineheight{1.25}\smash{\begin{tabular}[t]{l}$\kappa_1$\end{tabular}}}}%
    \put(0.01995366,0.25251568){\color[rgb]{0,0,0}\makebox(0,0)[lt]{\lineheight{1.25}\smash{\begin{tabular}[t]{l}$\kappa_2$\end{tabular}}}}%
    \put(0,0){\includegraphics[width=\unitlength,page=2]{objectivecandidates_b.pdf}}%
    \put(0.63485329,0.57297165){\color[rgb]{0,0,0}\makebox(0,0)[lt]{\lineheight{1.25}\smash{\begin{tabular}[t]{l}$p(\bm{\kappa} | \varphi^*)$\end{tabular}}}}%
    \put(0.53622736,0.20647157){\color[rgb]{0,0,0}\makebox(0,0)[lt]{\lineheight{1.25}\smash{\begin{tabular}[t]{l}\textcolor{customblueinksc1}{$\bm{\kappa}_{target}$}\end{tabular}}}}%
    \put(0,0){\includegraphics[width=\unitlength,page=3]{objectivecandidates_b.pdf}}%
  \end{picture}%
\endgroup%
}
\subcaption{Target value $\hp_{target}$}\label{fig:objectivetypes-b}
\end{minipage}%
\begin{minipage}{0.3\textwidth}
  \centering
\resizebox{0.90\linewidth}{!}{
\begingroup%
  \makeatletter%
  \providecommand\color[2][]{%
    \errmessage{(Inkscape) Color is used for the text in Inkscape, but the package 'color.sty' is not loaded}%
    \renewcommand\color[2][]{}%
  }%
  \providecommand\transparent[1]{%
    \errmessage{(Inkscape) Transparency is used (non-zero) for the text in Inkscape, but the package 'transparent.sty' is not loaded}%
    \renewcommand\transparent[1]{}%
  }%
  \providecommand\rotatebox[2]{#2}%
  \newcommand*\fsize{\dimexpr\f@size pt\relax}%
  \newcommand*\lineheight[1]{\fontsize{\fsize}{#1\fsize}\selectfont}%
  \ifx\svgwidth\undefined%
    \setlength{\unitlength}{174.62322831bp}%
    \ifx\svgscale\undefined%
      \relax%
    \else%
      \setlength{\unitlength}{\unitlength * \real{\svgscale}}%
    \fi%
  \else%
    \setlength{\unitlength}{\svgwidth}%
  \fi%
  \global\let\svgwidth\undefined%
  \global\let\svgscale\undefined%
  \makeatother%
  \begin{picture}(1,0.69330564)%
    \lineheight{1}%
    \setlength\tabcolsep{0pt}%
    \put(0,0){\includegraphics[width=\unitlength,page=1]{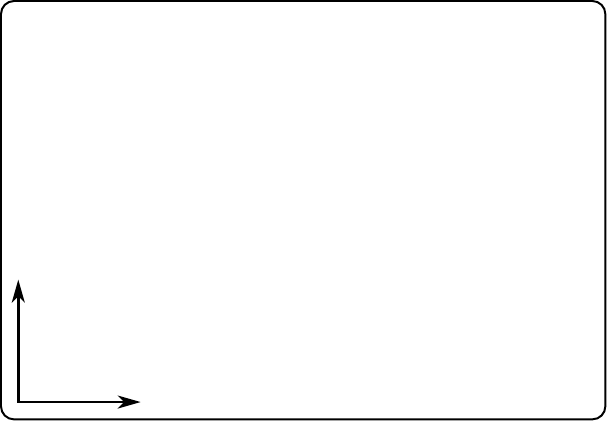}}%
    \put(0.24850682,0.02319553){\color[rgb]{0,0,0}\makebox(0,0)[lt]{\lineheight{1.25}\smash{\begin{tabular}[t]{l}$\kappa_1$\end{tabular}}}}%
    \put(0.01995366,0.25251568){\color[rgb]{0,0,0}\makebox(0,0)[lt]{\lineheight{1.25}\smash{\begin{tabular}[t]{l}$\kappa_2$\end{tabular}}}}%
    \put(0,0){\includegraphics[width=\unitlength,page=2]{objectivecandidates_c.pdf}}%
    \put(0.63485329,0.57297165){\color[rgb]{0,0,0}\makebox(0,0)[lt]{\lineheight{1.25}\smash{\begin{tabular}[t]{l}$p(\bm{\kappa} | \varphi^*)$\end{tabular}}}}%
    \put(0.53622736,0.20647157){\color[rgb]{0,0,0}\makebox(0,0)[lt]{\lineheight{1.25}\smash{\begin{tabular}[t]{l}\textcolor{customblueinksc1}{$p_{target} \left( \bm{\kappa} \right)$}\end{tabular}}}}%
    \put(0,0){\includegraphics[width=\unitlength,page=3]{objectivecandidates_c.pdf}}%
  \end{picture}%
\endgroup%
}
\subcaption{Target distribution $p_{target} \left( \hp \right)$}\label{fig:objectivetypes-c}
\end{minipage}
\caption{\textbf{Illustration of various materials design objectives.} Different optimization objectives with  respect to the density  $p \left( \hp \middle| \varphi \right)$ that expresses the likelihood of  property values $\hp$ for given processing conditions $\rfp$. We illustrate the following cases: (a)  we seek to maximize the probability that the material properties $\hp$ fall within a target domain $\mathcal{K}$.
(b) we seek to minimize the mean deviation of the properties $\hp$  from a target value  $\hp_{target}$. (c)  we seek to minimize the deviation between  $p \left( \hp \middle| \varphi \right)$ and a target probability density $p_{target} \left( \hp \right)$ on the material properties. } 
\label{fig:objectivetypes}
\end{figure*}

With these definitions in hand, we proceed to define two closely related optimization problems (O1) \& (O2) that we would like to address.
For the first optimization problem (O1) we make use of a utility function $u(\hp) \ge 0 $\footnote{Negative-valued utility functions can be employed as long as they are bounded from below i.e. $u(\hp) \ge u_0>-\infty$ in which case  $u(\hp) -u_0$ should be used   in place of $u(\hp)$}, specific examples of which are provided in the sequel.
Due to the uncertainty in $\hp$ we consider the {\em expected utility $U(\varphi)$} which is defined as: 
\be
U_1(\varphi) = \mathbb{E}_{p \left( \bm{x} \middle| \varphi \right)} \left[ \int u(\hp) ~p(\hp|\bx)~d\hp \right]
\label{eq:obj1}
\ee
(where $\mathbb{E}_{p \left( \bm{x} \middle| \varphi \right)}[.]$ implies an expectation with respect to $p\left( \bm{x} \middle| \varphi \right)$) and seek the processing parameters $\varphi$ that maximize it, i.e.:
\be
 \textrm{(O1):} \qquad \rfp^* = \arg \max_{\rfp} ~ U \left( \rfp \right) 
 \label{eq:o1}
 \ee
Consider for example  $u(\hp)$ being the indicator function  $\mathbb{I}_{\mathcal{K}} \left(  \hp \right)$ of some target domain $\mathcal{K}$, defining the desired range of property values (Figure \ref{fig:objectivetypes-a}). In this case, solving (O1) above will lead to the value of $\rfp$ that {\em maximizes} the probability that the resulting material will have properties in the target domain $\mathcal{K}$, i.e. $U \left( \varphi \right) = p \left( \hp \in \mathcal{K} \middle| \varphi \right)$. 
Similar {\em probabilistic} objectives have been proposed for several other materials' classes and models (e.g.  \cite{ikebata_bayesian_2017}).
Another possibility of potential practical interest involves $u(\hp)=e^{-\tau || \hp-\hp_{target}||^2}$\footnote{where $\tau$ is merely a scaling parameter} in which case solving (O1) leads to  the material  with properties which, on average, are closest to the prescribed target $\hp_{target}$ (Figure \ref{fig:objectivetypes-b}).  

The second problem we consider involves prescribing a target density $p_{target}(\hp)$ on the material properties and seeking the $\rfp$ that leads to a marginal density of properties $p \left( \hp \middle| \varphi \right) = \mathbb{E}_{p \left( \bm{x} \middle| \varphi \right)} \left[ p \left( \hp \middle| \bm{x} \right) \right]$  that is as close as possible to the target density (Figure \ref{fig:objectivetypes-c}). While there are several distance measures in the space of densities, we employ here the Kullback-Leibler divergence $KL(p_{target}(\hp) || p(\hp | \rfp))$, the minimization of which is equivalent to:
\be
\begin{array}{ll}
\textrm{(O2):} \qquad \rfp^*  & = \arg \max_{\rfp} U_2(\rfp)  \\
\textrm{ where } U_2(\rfp) &= \int p_{target}(\hp) \log p(\hp |\rfp) ~d\hp 
\end{array}
\label{eq:o2}
\ee
The aforementioned  objective resembles the one employed in \cite{tran_solving_2021}, but rather than finding a density on the microstructure (or features thereof) that leads to a close match of $p_{target}(\hp)$, we identify the processing variables $\rfp$ that do so.

We note that  both problems are considerably more challenging than deterministic counterparts, as in both cases the objectives involve expectations with respect to the high-dimensional vector(s) $\bx$ (and potentially $\hp$), representing the microstructure (and their effective properties). Additionally in the case of (O2), the analytically intractable density  $p(\hp |\rfp)$ appears explicitly in the objective. While one might argue that a brute-force Monte Carlo approach with a sufficiently large number of samples would suffice to carry out the aforementioned integrations, we note that propagating the uncertainty from $\bx$ to the properties $\hp$ would also  require commensurate solves of the expensive structure-property model which would need to be repeated for various $\rfp$-values. To overcome  challenges associated with the structure-property link,  we make use of the surrogate model i.e. instead of the true $p(\hp| \bx)$ in the expressions above we make use of $p_{\mathcal{M}}(\hp| \bx, \mathcal{D})$, based on a surrogate model $\mathcal{M}$.
The reformulated objectives are denoted with $U_{1,\mathcal{M}}^{\mathcal{D}}$ and $U_{2,\mathcal{M}}^{\mathcal{D}}$ and the specifics of the probabilistic surrogate, as well as the solution strategy proposed are explained in the sequel.

\subsubsection*{Expectation-Maximization and Stochastic Variational Inference}
\label{sec:emsvi}

We  present the proposed algorithm for the solution of (O1) and discuss the requisite changes for (O2) afterward.
Due to the  intractability of the objective function $U_{1,\mathcal{M}}^{\mathcal{D}}(\rfp)$  (see \refeqp{eq:o1}) and its derivatives, we employ the Expectation-Maximization scheme \cite{dempster_maximum_1977} which is based on 
 the so-called Evidence Lower BOund (ELBO) $\mathcal{F}$:
\be
\begin{array}{ll}
\log U_{1,\mathcal{M}}^{\mathcal{D}}(\rfp)  & =   \log \int u(\hp) ~p_{\mathcal{M}}(\hp|\bx, \mathcal{D}) ~p(\bx| \varphi)~d\hp~d\bx \\
&\geq \mathbb{E}_{q(\bx,\hp)} \left[ \log \cfrac{u(\hp) ~p_{\mathcal{M}}(\hp|\bx,\mathcal{D}) ~p(\bx| \varphi) }{q(\bx,\hp)} \right] \\
&= \mathcal{F} \left(  q \left( \hp, \bm{x} \right), \varphi \right)
\end{array}
\label{eq:elbo1}
\ee
where $\mathbb{E}_{q(\bx,\hp)}[.]$ denote the  expectation with respect to the auxiliary density $q(\bx,\hp)$. The algorithm alternates between maximizing $\mathcal{F}$ with respect to the density $q(\bx,\hp)$ while $\rfp$ is fixed (E-step) and maximizing with respect to $\rfp$ (M-step) while $q(\bx,\hp)$ is fixed.  We employ a Variational-Bayesian relaxation \cite{beal_variational_2003}, in short  VB-EM,  according to which instead of the optimal $q$ we consider a  family $\mathcal{Q}_{\bs{\xi}}$ of densities parameterized by $\bs{\xi}$ and in the E-step maximize $\mathcal{F}$ with respect to $\bs{\xi}$. This, as well as the the maximization with respect to $\rfp$ in the M-step,  is done by using stochastic gradient ascent  where the associated derivatives are substituted by noisy Monte Carlo estimates (i.e., Stochastic Variational Inference \cite{hoffman_stochastic_2013}).
The particulars of $\bs{\xi}$ as well as of the E- and M-steps are discussed in the Supplementary Information. We illustrate the basic, numerical  steps in the inner-loop of Algorithm \ref{Alg:EM} (the algorithm starts from an initial, typically random, guess of $\bm{\xi}$ and $\varphi$).
Colloquially, the VB-EM iterations  can be explained as follows: In the E-step and given the current estimate for $\varphi$, one averages over microstructures that are not only a  priori more probable according to $p(\bx |\varphi)$ but also achieve a higher score according to $u(\hp) ~ p_{\mathcal{M}} (\hp | \bx, \mathcal{D})$.  
Subsequently, in the M-step step, we update the optimization variables $\rfp$ on the basis of the average above (see Supplementary Information - VB-EM-Algorithm). 

The second objective,  $U_{ 2,\mathcal{M} }$ (\refeqp{eq:o2}) can be dealt with in a similar fashion. As it involves an integration over $\hp$ with respect to the target density $p_{target}\left( \vect{\kappa} \right)$, it can first be approximated using  $S$ Monte Carlo samples $\{\hp^{(s)}\}_{s=1}^S$ from the $p_{target}\left( \vect{\kappa} \right)$ and subsequently each of the terms in the sum can be lower-bounded as follows:

\be
\begin{array}{ll}
U_{2, \mathcal{M}}^{\mathcal{D}}(\rfp) &= \int p_{target}(\hp) \log p_{\mathcal{M}}(\hp |\rfp, \mathcal{D}) ~d\hp \\
& \approx \frac{1}{S} \sum_{s=1}^S \log p_{\mathcal{M}}(\hp^{(s)} |\rfp, \mathcal{D})  \\
& = \frac{1}{S} \sum_{s=1}^S \log \int p_{\mathcal{M}}(\hp^{(s)} | \bx , \mathcal{D})~p(\bx|\rfp)~d\bx \\
& \ge \frac{1}{S} \sum_{s=1}^S \mathbb{E}_{q^{(s)}(\bx)} \left[ \log \cfrac{ p_{\mathcal{M}}(\hp^{(s)} | \bx , \mathcal{D})~p(\bx|\rfp)}{q^{(s)}(\bx)} \right] \\
& = \frac{1}{S} \sum_{s=1}^S\mathcal{F}_s \left(  q^{(s)} \left( \bm{x} \right), \varphi \right)
\end{array}
\label{Eq:ELBO_O2}
\ee

In this case, the aforementioned stochastic variational inference tools will need to be applied for  updating  each $q^{(s)} \left( \vect{x} \right), s = 1, ..., S$ in the E-step, but the overall algorithm remains conceptually identical. We note that  incremental and partial versions where, e.g., a subset of the $q^{(s)}$ are updated with one or more steps of stochastic gradient ascent are possible \cite{neal1998view} and can lead to improved computational performance. \\

\subsubsection*{Probabilistic surrogate model}
\label{sec:surrogate}

In order to overcome the computational roadblock  imposed by the high-fidelity model employed in the structure-property link (i.e. $\hp(\bx)$ or $p(\hp |\bx)$), we substitute it with a data-driven surrogate which is trained on $N$ pairs of data 

\be
\mathcal{D} = \left\lbrace \vect{x}^{(n)}, \hp^{(n)} = \hp \left( \vect{x}^{(n)} \right) \right\rbrace_{n=1}^N
\ee
While such supervised machine learning problems have been studied extensively and a lot of the associated tools have found their way in materials applications \cite{kalidindi_bayesian_2019}, we note that their use in the context of the optimization problems presented requires significant adaptations. 

In particular and unlike canonical, data-centric applications which rely on the abundance of data (Big Data), we operate under a smallest-possible-data regime.
This is because in our setting training data arise from expensive simulations, and consequently we are interested in minimizing the number of data points which have to be generated.
The shortage of information generally leads to increased  predictive uncertainty which, rather than dismissing, we quantify by employing a {\em probabilistic surrogate} that yields a predictive density $p_{\mathcal{M}}(\hp| \bx,\mathcal{D})$ instead of mere point estimates.
More importantly though,  we note that the distribution of the inputs in $\mathcal{D}$, i.e. the microstructures $\bx$, changes drastically with $\rfp$ (Figure \ref{fig:overview}). As we do not know a priori the optimal $\rfp^*$, we cannot generate training data from $p(\bx|\rfp^*)$ and data-driven surrogates generally produce poor extrapolative, out-of-distribution predictions \cite{marcus_rebooting_2019}.  
It is clear therefore, that the selection of the training data, i.e. the microstructures-inputs $\bx^{(n)}$ for which we pay the price of computing the output-property of interest $\hp^{(n)}$, should be integrated with  the optimization algorithm in order to produce a sufficiently accurate surrogate
while keeping $N$ as small as possible. We defer a detailed discussion of this aspect for the next section, and will first present the particulars of the surrogate model employed. 

The probabilistic surrogate adopted for the purpose of driving the optimization algorithm makes use of a multivariate Gaussian likelihood, i.e.  $\hp | \bx \sim \mathcal{N} \left( \bs{m}_{\bt}(\bx), \bs{S}_{\bt}(\bx)\right)$, where the mean $\bs{m}_{\bt}(\bx)$ and covariance $\bs{S}_{\bt}(\bx)$ are modeled with a convolutional neural network (CNN) (see Figure \ref{fig:CNN_architecture} and Suppl. Information - Probabilistic Surrogate), with $\bt$ denoting the associated parameters. The use of such convolutional neural networks has been previously proposed for property prediction in binary media in e.g., \cite{yang2018deep, cecen2018material}. Point estimates $\bt_{\mathcal{D}}$ of the parameters are obtained with the help of training data $\mathcal{D}$ by maximizing the corresponding likelihood $p \left( \mathcal{D} \middle| \bm{\theta} \right)$. On the basis of these estimates, the predictive density (i.e for a new input-microstructure $\bx$) of the surrogate follows as $p_{\mathcal{M}}(\hp| \bx, \mathcal{D})=\mathcal{N} \left( \bs{m}_{\bt_{\mathcal{D}}}(\bx), \bs{S}_{\bt_{\mathcal{D}}}(\bx)\right)$. We emphasize the dependence of the probabilistic surrogate on the dataset $\mathcal{D}$, for which we will discuss an adaptive acquisition strategy in the following section.
While the results obtained are based on this particular architecture of the surrogate, the methodological framework  proposed can accommodate any probabilistic  surrogate  and integrate its predictive uncertainty in the optimization procedure.

\subsubsection*{Active Learning}
\label{sec:al}
Active learning refers to a family of methods whose goal is to improve learning accuracy and efficiency  by selecting particularly salient training data \cite{tong_active_2001}. This is especially relevant in our applications as data acquisition is de facto the most computationally expensive component.
The basis of all such methods is to progressively enrich the training dataset by scoring candidate inputs (i.e. microstructures $\bx$ in our case) based on their {\em expected informativeness} \cite{mackay_information-based_1992}. 
The latter can be quantified with a so-called acquisition function $\alpha(\bx)$
 , for which many different forms have been proposed (depending on the specific setting). We note though that in most cases in the literature, acquisition functions associated with the predictive accuracy of the supervised learning model have been employed, which in our formulation  translates to the accuracy of our surrogate in  predicting the properties $\hp$ for an  input microstructure. In this regard, since a  measure of the predictive uncertainty is  the covariance $\bs{S}_{\bt_{\mathcal{D}}}(\bx)$ above, one might define e.g. $\alpha(\bx)=trace\left( \bs{S}_{\bt_{\mathcal{D}}}(\bx) \right)$.  Alternative acquisition functions have been proposed in the context of Bayesian Optimization problems which as explained in the introduction exhibit significant  differences with ours \cite{frazier_bayesian_2016}. While it is true that a perfect surrogate, i.e., when $p(\hp |\bx) = p_{\mathcal{M}}(\hp| \bx, \mathcal{D}) ~\forall \bx$, would yield the exact optimum, imperfect surrogates could be useful as long as they can correctly guide the search in the $\rfp$-space and lead to optimal value of (O1) or (O2). Hence an accurate surrogate for $\rfp-$values (and corresponding microstructures $\bx$) far away from the optimum is not necessary. The difficulty of course is that we do not know a priori where the optimum lies or not and a surrogate trained on microstructures drawn from $p(\bx | \rfp^{(0)})$ where $\rfp^{(0)}$ is the initial guess in the optimization scheme (see Algorithm \ref{Alg:EM}), will generally perform poorly at other $\rfp$'s.

The acquisition function that we propose incorporates the optimization objectives. In particular for the (O1) problem (\refeqp{eq:obj1}) we propose:
\be
\alpha(\bx)=\text{Var}_{p_{\mathcal{M}}(\hp | \bx, \mathcal{D}) }\left[ u(\hp) \right]
\label{eq:acquisition}
\ee
In this particular form, $\alpha$ scores each microstructure $\bx$ in terms of  the predictive uncertainty in the utility $u$ (the expected value of which we seek to maximize) due to the predictive density of the surrogate. 
In the case discussed earlier where $u(\hp)= \mathbb{I}_{\mathcal{K}} \left(  \hp \right)$ (and $U_{1}(\rfp) = Pr \left( \hp \in \mathcal{K} \middle| \varphi \right)$), the acquisition function reduces to the variance of the event $\hp \in \mathcal{K}$. This suggests that the acquisition function yields the largest scores for microstructures for which the surrogate is most uncertain whether their corresponding properties belong in the target domain $\mathcal{K}$

In terms of the overall procedure, we propose an outer loop, within which the VB-EM-based optimization is embedded, such  that if $\mathcal{D}^{(l)}$ denotes the training dataset at iteration $l$, $p_{\mathcal{M}}(\hp |\bx,\mathcal{D}^{(l)})$ the corresponding  predictive density of the surrogate,  $q^{(l)}(\bx)$ the marginal variational density found in the last E-step and $\rfp^{(l)}$ the optimum found in the last M-step, we augment the training dataset as follows (see Algorithm \ref{Alg:EM}):
\bi
\item we randomly generate a pool of candidate microstructures $\{\bx^{(l,n)}\}_{n=1}^{N_{pool}}$ from  $q^{(l)}(\bx)$ and select a subset of  $N_{add}<N_{pool}$ microstructures which yield the highest values of the acquisition function $\alpha(\bx^{(l,n)})$
\item  We solve the high-fidelity model for the aforementioned $N_{add}$ microstructures and construct a new training dataset $D_{add}^{(l)}$ which we add to  $\mathcal{D}^{(l)}$ in order to form $\mathcal{D}^{(l+1)}=\mathcal{D}^{(l)} \cup D_{add}^{(l)}$.  We retrain \footnote{The retraining could be avoided by making use of online learning \cite{sahoo2017online}, which does not rely on an a-priori fixed dataset and can deal with incrementally arriving (or streamed) data} the surrogate based on $\mathcal{D}^{(l+1)}$, i.e. we compute  $p_{\mathcal{M}}(\hp|\bx, \mathcal{D}^{(l+1)})$,  and restart the VB-EM-based optimization algorithm with the now updated surrogate. 
\ei

For the (O2) problem above, we propose to select microstructures that yield the highest predictive log-score on the sample representation $\{ \hp^{(s)} \}_{s=1}^S $ of the target distribution, i.e.

\begin{align}
    \alpha \left( \bm{x} \right) = \frac{1}{S} \sum\limits_{s=1}^S \log p_{\mathcal{M}} \left( \hp^{(s)} \middle| \bm{x} \right) \qquad \hp^{(s)} \sim p_{target} \left( \hp \right)
    \label{Eq:A1}
\end{align}

\section*{Results \& Discussion}
\label{sec:results}

In the following, we present two applications of the proposed framework for (O1)- and (O2)-type optimization problems.  We first elaborate on the specific choices for the process parameters $\rfp$, the random microstructures $\bx$ and their properties $\hp$ as well as the associated PSP links.

\textbf{Process $\rfp$ - Microstructure $\bx$:} In all numerical illustrations we consider statistically homogeneous,  binary (two-phase) microstructures which upon spatial discretization (on a uniform, two-dimensional $N_p\times N_p$ grid with $N_p=64$) are represented by a vector $\bx \in \{0,1 \}^{4096}$. The binary microstructures are modeled by means of a thresholded zero-mean, unit-variance  Gaussian field \cite{teubner_level_1991,roberts_transport_1995}. If the vector $\bx_g$ denotes the discretized version of the latter (on the same grid), then the value at each pixel $i$ is given by $x_i=H(x_{g,i}-x_0)$ where   $H(\cdot)$ denotes the Heaviside function and $x_0$ the cutoff threshold. which determines the volume fractions of the resulting binary field. We parameterize with $\rfp$ the  spectral density function (SDF) of the underlying Gaussian field  (i.e. the Fourier transform of its autocovariance) using a combination  of radial basis functions (RBFs, see Supplementary Information-Process-Structure linkage) which automatically ensures the non-negativity of the resulting SDF.
The constraint of unit variance is enforced using a softmax transformation. The density  $p(\bx |\rfp)$ implicitly defined above affords great flexibility in the forms of the resulting binary medium (as can be seen in the ensuing illustrations) which increases as the dimension of $\rfp$ does. 
Figure \ref{fig:overview} illustrates how different values of the process parameters $\varphi$ can lead to profound changes in the microstructures (and correspondingly, their effective physical properties $\hp$). While the parameters $\rfp$ selected do not have explicit physical meaning, they can be linked to actual processing variables given appropriate data. 
Naturally, not all binary media can be represented by this model and a more flexible $p(\bx |\rfp)$ could be employed with small modifications in the overall algorithm \cite{koutsourelakis_probabilistic_2006,bostanabad_stochastic_2016,cang_microstructure_2017}.

\textbf{Microstructure $\bx$ - Properties $\hp$ }
In this study we consider a two-dimensional, representative volume element (RVE) $\Omega_{\text{RVE}}=[0,1]^2$ and assume each of  the two phases are isotropic, linear elastic in terms of their  mechanical response and are characterized by isotropic,  linear conductivity tensors in terms of their thermal response.  We denote  with $\mathbb{C}$ the fourth-order elasticity tensor and with $\mathbf{a}$ the second order conductivity tensor which are also binary (vector and tensor) fields. The vector $\hp$ consists of various combinations of  macroscopic, effective (apparent), mechanical or thermal properties of the RVE which we denote by $\mathbb{C}^{\text{eff}}$ and $\mathbf{a}^{\text{eff}}$, respectively. The effective properties for each microstructure occupying $\Omega_{\text{RVE}}$ were computed using finite element simulations and Hill's averaging theorem \cite{miehe2002computational, hill1972constitutive} (see Supplementary Information: Structure-Property linkage). We assumed a contrast ratio of $50$ in the properties of the two phases, i.e., $E_1 / E_0=50$ (where $E_0,E_1$ are the elastic moduli of phases $0$ and $1$)\footnote{Poisson's ratio was $\nu=0.3$ for both phases} and $a_1 / a_0=50$ (where $a_0,a_1$ are the conductivities of phases $0$ and $1$). In the following plots, phase 1 is  always  shown as white and phase 0 always as black. We note that the dependence of effective properties  on (low-dimensional) microstructural features (analogous to $\rfp$) has been considered, in e.g. \cite{saheli_microstructure_2004,fullwood_microstructure_2010} but the random variability in these properties has been ignored either by considering very large RVEs or by averaging over several of them.  We emphasize finally that the framework proposed can accommodate any high-fidelity model for the structure-property link as this is merely used as a generator of training data $\mathcal{D}$.


\subsubsection*{Case 1:  Target domain of multi-physics properties (O1) }

\begin{figure*}[hbtp]
\centering 
\begin{center}

    \includegraphics[height=6.1cm, width=0.975\linewidth]{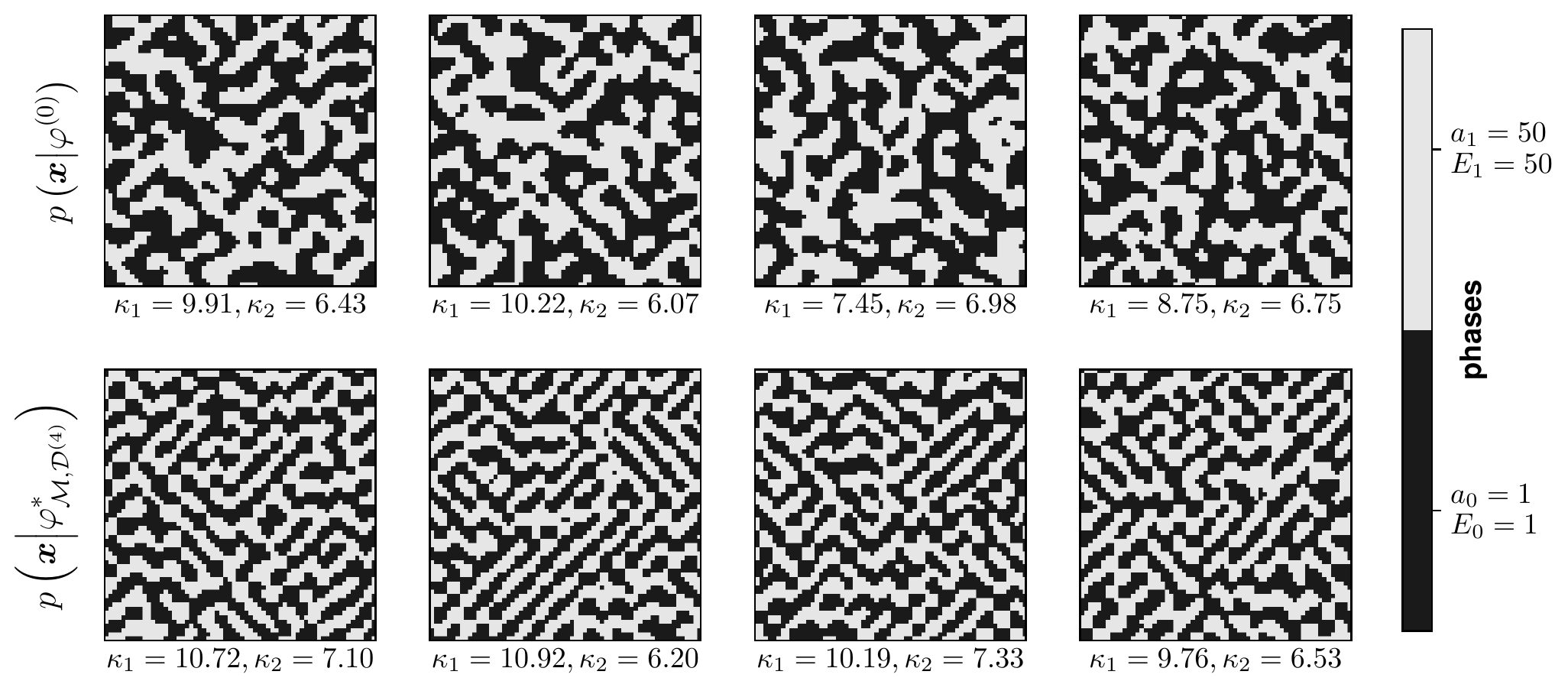}
	\caption{ \textbf{Case 1: Optimal random microstructures.} \textit{(Tow row)} Samples of microstructures drawn from $p(\bx|\rfp)$ for the initial guess $\varphi^{(0)}$ of processing variables, \textit{(Bottom row)} Samples of microstructures drawn from $p(\bx|\rfp)$ for the optimal value  $\varphi^{*}_{\mathcal{M}, \mathcal{D}^{(L)}}$ of processing variables  which maximize the probability that the corresponding material properties will fall in the target domain  $\mathcal{K}=\left[ 8.5, ~ 11.0 \right] \times   \left[ 6.75, ~  9.0 \right]$ (\refeqp{Eq:target_domain_values}). Underneath each microstructure, the thermal $\kappa_1$ and mechanical $\kappa_2$ properties of interest (\refeqp{eq:prop1}) are reported.}
	\label{fig:multiphysics_microstructures}
	\vspace{0.15cm}
        
	\includegraphics[height=5cm,width=0.91\linewidth]{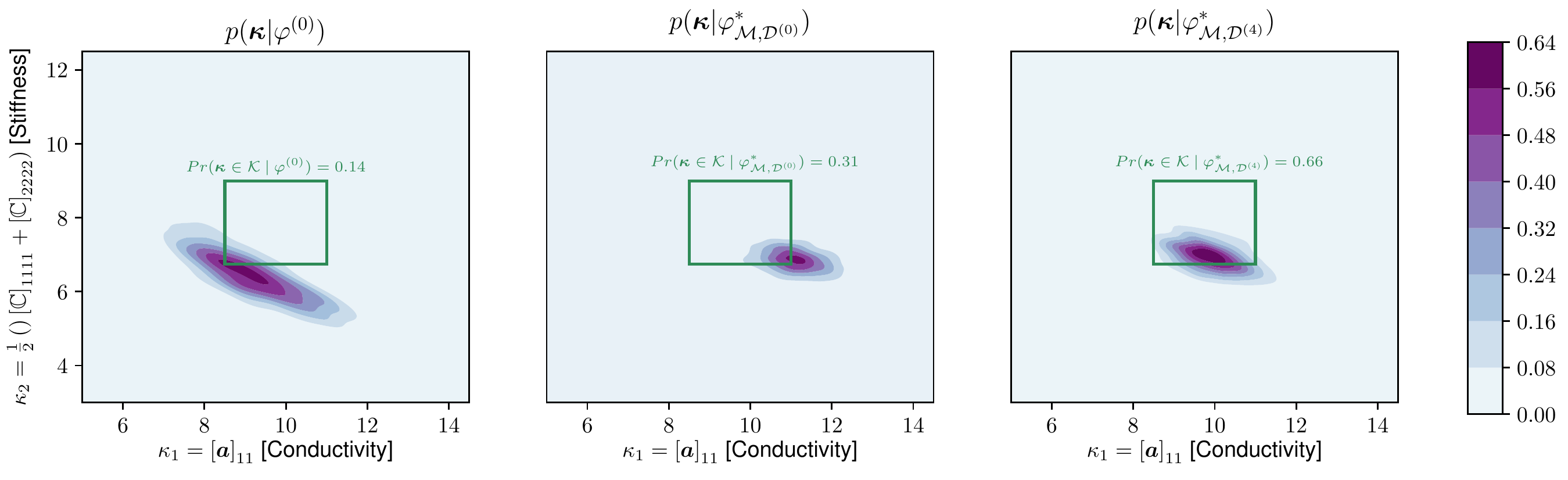}
	\caption{\textbf{Case 1: Evolution of the process-property density during optimization.}
	The actual process-property density $p \left( \hp \middle| \varphi \right)$ was estimated using \tmp{$1024$} Monte Carlo samples, the high-fidelity structure-property model (see Supplementary Information: Structure-Property linkage) and for three values of the process parameters $\rfp$. \textit{(Left)} for the initial guess
	$\varphi^{(0)}$, \textit{(Middle)} for the optimal $\varphi$ as obtained using the initial training dataset $\mathcal{D}^{(0)}$ 
	and without adaptive learning, \textit{(Right)} for the optimal $\varphi$  obtained with the augmented training dataset $\mathcal{D}^{(4)}$  identified by the active learning scheme proposed and as detailed in the text.
	The target domain $\mathcal{K}$ (\refeqp{Eq:target_domain_values}) is drawn with  a green rectangle and the colorbar indicates the value of the density $p \left( \hp \middle| \varphi \right)$.}
	\label{fig:multiphysics_distribution2d}
	\vspace{0.15cm}

\setlength{\figH}{6.0cm}
\setlength{\figW}{0.50\linewidth}
\begin{tikzpicture}


\begin{groupplot}[group style={group size=2 by 1, horizontal sep=1.5cm}]
\nextgroupplot[
height=\figH,
legend cell align={left},
legend style={
  fill opacity=0.8,
  draw opacity=1,
  text opacity=1,
  at={(0.91,0.5)},
  anchor=east,
  draw=white!80!black
},
tick align=outside,
tick pos=left,
title={\(\displaystyle \mathcal{S} = Pr \left( \kappa \in \mathcal{K} | \varphi^*_{\mathcal{M}, \mathcal{D}^{(l)}} \right)\)},
width=\figW,
x grid style={white!69.0196078431373!black},
xlabel={N : labeled data (microstructures)},
xmin=1843.2, xmax=6348.8,
xtick style={color=black},
y grid style={white!69.0196078431373!black},
ylabel={achieved metric \(\displaystyle \mathcal{S}\)},
ymin=0.42861328125, ymax=0.62412109375,
ytick style={color=black}
]
\addplot [thick, color0, mark=*, mark size=3, mark options={solid}]
table {%
2048 0.45703125
3072 0.552734375
4096 0.59375
5120 0.611328125
6144 0.615234375
};
\addlegendentry{active learning}
\addplot [thick, color1, mark=*, mark size=3, mark options={solid}]
table {%
2048 0.4375
3072 0.453125
4096 0.451171875
5120 0.46484375
6144 0.45703125
};
\addlegendentry{random (baseline)}

\nextgroupplot[
height=\figH,
legend cell align={left},
legend style={
  fill opacity=0.8,
  draw opacity=1,
  text opacity=1,
  at={(0.91,0.5)},
  anchor=east,
  draw=white!80!black
},
tick align=outside,
tick pos=left,
title={marginal \(\displaystyle p(\kappa_1)\)},
width=\figW,
x grid style={white!69.0196078431373!black},
xlabel={effective property \(\displaystyle \kappa_1\)},
xmin=5.78063516616821, xmax=13.6995093345642,
xtick style={color=black},
y grid style={white!69.0196078431373!black},
ylabel={marginal \(\displaystyle p ( \kappa_1 )\)},
ymin=-0.0296439955909816, ymax=0.622523907837888,
ytick style={color=black}
]
\addplot [line width=1pt, black]
table {%
6.7114405632019 4.7733105744394e-10
6.84670831719223 2.69010420912383e-08
6.98197607118256 8.5823526042954e-07
7.11724382517289 1.5517287222155e-05
7.25251157916322 0.000159198059858489
7.38777933315355 0.000928317671399684
7.52304708714388 0.00308963681000967
7.65831484113421 0.00599023569908531
7.79358259512454 0.00761314839677616
7.92885034911487 0.0099118119782084
8.06411810310519 0.0185120150561497
8.19938585709552 0.0337124546938584
8.33465361108585 0.0518236314496312
8.46992136507618 0.0757421535214548
8.60518911906651 0.111422982055329
8.74045687305684 0.15770273938273
8.87572462704717 0.20946540748438
9.0109923810375 0.26411119067236
9.14626013502783 0.326555027060036
9.28152788901816 0.399453417688136
9.41679564300848 0.468386845221105
9.55206339699881 0.516782857409704
9.68733115098914 0.537941607071629
9.82259890497947 0.534044790213766
9.9578666589698 0.520475240572273
10.0931344129601 0.506415490777553
10.2284021669505 0.477118527945203
10.3636699209408 0.428375436254661
10.4989376749311 0.376735810188734
10.6342054289214 0.324349752113874
10.7694731829118 0.26373561451051
10.9047409369021 0.203919731595708
11.0400086908924 0.15796211503227
11.1752764448828 0.124321540251306
11.3105441988731 0.0945794557989135
11.4458119528634 0.0672769298876889
11.5810797068537 0.0442122375268429
11.7163474608441 0.0268107806628293
11.8516152148344 0.0162030316774223
11.9868829688247 0.0103949149474797
12.1221507228151 0.00687751288089527
12.2574184768054 0.00467235544481007
12.3926862307957 0.00297612914450863
12.5279539847861 0.00143425263026709
12.6632217387764 0.000442493462345807
12.7984894927667 8.07324872876325e-05
12.933757246757 8.43896585199817e-06
13.0690250007474 4.99269530244471e-07
13.2042927547377 1.66397344579067e-08
13.339560508728 3.11839694740364e-10
};
\addlegendentry{$p ( \kappa_1 | \varphi^*_{\mathcal{M}, \mathcal{D}^{(4)}})$}
\addplot [line width=1pt, color0, dashed]
table {%
6.14058399200439 2.58799914212462e-11
6.26693996118039 2.22414685353126e-09
6.3932959303564 1.06674100038554e-07
6.5196518995324 2.87596437565331e-06
6.6460078687084 4.41996685548607e-05
6.7723638378844 0.000397956503768037
6.8987198070604 0.0022090402381534
7.0250757762364 0.00820669858391773
7.1514317454124 0.0224145540725715
7.2777877145884 0.0476876876016896
7.4041436837644 0.0805227291399311
7.5304996529404 0.112736439698516
7.6568556221164 0.145492230254396
7.7832115912924 0.192054287358125
7.9095675604684 0.259421209660447
8.0359235296444 0.33942421792354
8.1622794988204 0.420047798814377
8.2886354679964 0.490494136207777
8.4149914371724 0.53669002259656
8.5413474063484 0.55513799174385
8.6677033755244 0.568633221231143
8.7940593447004 0.588008263917241
8.9204153138764 0.587475553022072
9.04677128305241 0.551948291850076
9.17312725222841 0.488130963247949
9.29948322140441 0.4059939306235
9.42583919058041 0.329374543358819
9.55219515975641 0.276201688273453
9.67855112893241 0.234996267673073
9.80490709810841 0.189411121599947
9.93126306728441 0.142492804923342
10.0576190364604 0.105131608912347
10.1839750056364 0.0784331233792659
10.3103309748124 0.0569040773989649
10.4366869439884 0.0383849589161609
10.5630429131644 0.0233189757977071
10.6893988823404 0.0123101778348799
10.8157548515164 0.00595710609545044
10.9421108206924 0.00351637772804506
11.0684667898684 0.00348315592448223
11.1948227590444 0.00404219062841092
11.3211787282204 0.00366492780723271
11.4475346973964 0.00225773523425439
11.5738906665724 0.000869740655860387
11.7002466357484 0.000197030420860808
11.8266026049244 2.53441906824754e-05
11.9529585741004 1.82064982144246e-06
12.0793145432764 7.25226126180888e-08
12.2056705124524 1.59710752576751e-09
12.3320264816284 1.94216081601339e-11
};
\addlegendentry{$p ( \kappa_1 | \varphi^*_{\mathcal{M},\mathcal{D}^{(4)}}, \mathcal{D}^{(0)} )$}
\addplot [line width=1pt, color1, dashed]
table {%
7.05646419525146 2.69140144591905e-11
7.17247571750563 1.68520044951133e-09
7.2884872397598 6.43210347695593e-08
7.40449876201396 1.50214016877079e-06
7.52051028426813 2.15898807170444e-05
7.63652180652229 0.00019271820501938
7.75253332877646 0.00108400791277839
7.86854485103062 0.00393826077298948
7.98455637328479 0.00968991713530955
8.10056789553895 0.0177969556686381
8.21657941779312 0.0285928754714748
8.33259094004728 0.0449790640151531
8.44860246230145 0.0672119712156932
8.56461398455561 0.0916595259718922
8.68062550680978 0.116856755171475
8.79663702906394 0.144425463882016
8.91264855131811 0.180178460372583
9.02866007357228 0.232289749583684
9.14467159582644 0.297027535555809
9.26068311808061 0.355672148975182
9.37669464033477 0.395388399457834
9.49270616258894 0.428939831969494
9.6087176848431 0.478697135685418
9.72472920709727 0.538665266097737
9.84074072935143 0.579625680499774
9.9567522516056 0.592879912227485
10.0727637738598 0.591995635937088
10.1887752961139 0.578891608446188
10.3047868183681 0.5463407638353
10.4207983406223 0.494904954311405
10.5368098628764 0.4275890683336
10.6528213851306 0.350854620376072
10.7688329073848 0.277592083538613
10.8848444296389 0.214474879661415
11.0008559518931 0.160129470636154
11.1168674741473 0.115707024730995
11.2328789964014 0.0825082269612673
11.3488905186556 0.0586095175917122
11.4649020409097 0.0422916050865017
11.5809135631639 0.0311704780981449
11.6969250854181 0.0217351083541699
11.8129366076722 0.0124369732179994
11.9289481299264 0.00511986044181221
12.0449596521806 0.00139662383453436
12.1609711744347 0.000242157477153935
12.2769826966889 2.61733700912948e-05
12.3929942189431 1.7473435504132e-06
12.5090057411972 7.17173679910134e-08
12.6250172634514 1.80473335963698e-09
12.7410287857056 2.77923563273192e-11
};
\addlegendentry{$p ( \kappa_1 | \varphi^*_{\mathcal{M},\mathcal{D}^{(4)}}, \mathcal{D}^{(4)} )$}
\end{groupplot}

\end{tikzpicture}
\caption{\textbf{Case 1: Assessment of active learning approach.} \textit{(Left:)} The probability we seek to maximize with respect to $\rfp$, i.e. $Pr \left( \hp \in \mathcal{K} \middle| \varphi \right)$ is plotted as a function of the size $N$ of the training dataset (i.e. the number of simulations of the high-fidelity model). The lines depict the medians obtained over $80$ independent runs of each algorithm. The red line corresponds to the results obtained without adaptive learning and the blue with adaptive learning.
\textit{(Right:)} for the the optimal $\rfpe{\mathcal{M}, \mathcal{D}^{(4)}}{*}$ identified using active learning,  we compare the  actual process-property density $p(\kappa_1| \rfp)$ (black line - estimated with $1024$ Monte Carlo samples and the high-fidelity model) with the one predicted by the surrogate trained only on the initial dataset $\mathcal{D}^{(0)}$ (blue line) and with the one predicted by the surrogate trained on the augmented dataset $\mathcal{D}^{(4)}$ (red line).
}
   \label{fig:MultiphysicsHybridPlot}
    
\end{center}
\end{figure*}

In the following we will demonstrate the performance of the proposed formulation in an (O1)-type stochastic optimization problem, with regards to both thermal as well as mechanical  properties.
 Additionally, we will provide a systematic and quantitative assessment of the benefits of the active learning strategy proposed (as compared to randomized data generation).

We consider a combination of mechanical and thermal  properties of interest, namely: 
\be
\kappa_1=[ \bm{a}^{\text{eff}} ]_{11}, \qquad  \kappa_2= \frac{1}{2} \left( [ \mathbb{C}^{\text{eff}} ]_{1111} + [ \mathbb{C}^{\text{eff}} ]_{2222} \right)
\label{eq:prop1}
\ee
 i.e., $\hp \in \mathbb{R}_{+}^2$, and define the  target domain:
\be
\mathcal{K}=\left[ 8.5, ~ 11.0 \right] \times   \left[ 6.75, ~  9.0 \right]
\label{Eq:target_domain_values}
\ee
The utility function $u(\hp)= \mathbb{I}_{\mathcal{K}} \left( \hp \right)$  is  the (non-differentiable) indicator function of $\mathcal{K} \subset \mathbb{R}^2_+$ which implies that the objective of the optimization (type (O1) - see Figure \ref{fig:objectivetypes-a}) is to {\em find the $\rfp$ that maximizes the probability that the resulting microstructures have properties $\hp$ that lie in $\mathcal{K}$}. The two-phase microstructures have volume fraction $0.5$ and the parameters $\rfp \in \RR^{100}$ as well as $p(\bx|\rfp)$ were defined as discussed in the beginning of this section\footnote{In this case, the centers of the $100$ RBFs are fixed to  a uniform grid and the bandwidths were prescribed. Hence the $100$ entries of $\rfp$ correspond to the weights of the RBFs}.

With regards to the adaptive learning strategy (appearing as the outer loop in Algorithm \ref{Alg:EM}), we note: the initial training dataset $\mathcal{D}^{(0)}$ consists of \tmp{$N_0 = 2048$} data pairs. This is generated via ancestral sampling, i.e. we randomly draw samples  $\rfp$ from  $\mathcal{N} ( \bm{0}, \mat{I} )$ \footnote{The choice $\varphi \sim \mathcal{N} ( \bm{0}, \mat{I} )$ is not arbitrary, as - given the adopted parametrization - it envelopes all possible SDFs}  and conditionally on each $\varphi^{(n)}$ we sample $p ( \bm{x} | \varphi^{(n)} )$ to generate a microstructure. In each data acquisition step $l$, $N_{pool}=4096$ candidates were generated and a subset of $N_{add}=1024$ candidates was selected based on the acquisition function. Hence the size of the dataset increased by \tmp{$1024$} data pairs at each iteration $l$, with \tmp{$L=4$} data augmentation steps performed in total.

\setlength{\figH}{4.8cm}
\setlength{\figW}{0.90\linewidth}
\begin{figure*}[t!]
\centering
\begin{minipage}{0.45\textwidth}
  \centering
\begin{tikzpicture}


\begin{axis}[
height=\figH,
tick align=outside,
tick pos=left,
title={EM - algorithm},
width=\figW,
x grid style={white!69.0196078431373!black},
xlabel={EM-iteration},
xmin=-5.75, xmax=120.75,
xtick style={color=black},
y grid style={white!69.0196078431373!black},
ylabel={ELBO \(\displaystyle \mathcal{F}\)},
ymin=1895.98131713867, ymax=2264.85742797852,
ytick style={color=black}
]
\addplot [thick, color0]
table {%
0 1912.74841308594
1 2027.18811035156
2 2105.35546875
3 2160.60668945312
4 2193.75439453125
5 2214.8896484375
6 2228.62060546875
7 2235.212890625
8 2239.75
9 2242.9130859375
10 2244.52661132812
11 2245.953125
12 2246.51904296875
13 2247.0361328125
14 2247.138671875
15 2247.75219726562
16 2247.646484375
17 2247.54638671875
18 2248.09033203125
19 2247.90625
20 2247.88330078125
21 2247.6513671875
22 2247.837890625
23 2248.025390625
24 2247.50390625
25 2247.73364257812
26 2247.48828125
27 2247.3974609375
28 2247.64111328125
29 2247.95556640625
30 2247.53784179688
31 2247.42602539062
32 2247.416015625
33 2247.4990234375
34 2247.07421875
35 2247.30078125
36 2246.9765625
37 2246.876953125
38 2246.88989257812
39 2247.23974609375
40 2247.4111328125
41 2247.068359375
42 2247.25537109375
43 2247.05859375
44 2247.54956054688
45 2247.28979492188
46 2247.2626953125
47 2247.23046875
48 2247.66748046875
49 2247.14697265625
50 2247.28955078125
51 2247.302734375
52 2246.95654296875
53 2247.22143554688
54 2247.072265625
55 2247.01220703125
56 2247.41162109375
57 2247.37548828125
58 2247.20947265625
59 2247.1748046875
60 2247.11181640625
61 2247.30419921875
62 2247.1259765625
63 2247.08129882812
64 2247.02294921875
65 2247.65698242188
66 2247.21435546875
67 2246.84985351562
68 2247.43603515625
69 2246.95361328125
70 2246.75610351562
71 2247.44848632812
72 2247.02490234375
73 2247.16015625
74 2247.51904296875
75 2247.0576171875
76 2247.39013671875
77 2247.0771484375
78 2247.36376953125
79 2247.26293945312
80 2247.45043945312
81 2246.65893554688
82 2247.00756835938
83 2247.1708984375
84 2246.931640625
85 2247.38330078125
86 2247.22338867188
87 2247.53247070312
88 2247.08837890625
89 2246.81103515625
90 2247.27197265625
91 2247.44555664062
92 2247.23999023438
93 2246.99243164062
94 2247.55029296875
95 2246.92114257812
96 2247.26147460938
97 2246.6630859375
98 2247.0595703125
99 2247.2705078125
100 2247.21533203125
101 2247.19970703125
102 2247.06713867188
103 2246.8818359375
104 2247.13671875
105 2247.03881835938
106 2247.01171875
107 2246.83935546875
108 2247.46728515625
109 2247.18115234375
110 2247.26806640625
111 2246.86669921875
112 2246.9228515625
113 2246.96875
114 2247.24926757812
115 2247.00732421875
};
\end{axis}

\end{tikzpicture}
\subcaption{Convergence of ELBO for $l=0$}\label{fig:elbo_convergence}
\end{minipage}%
\begin{minipage}{0.45\textwidth}
  \centering
\begin{tikzpicture}


\begin{axis}[
height=\figH,
tick align=outside,
tick pos=left,
title={Active Learning | \(\displaystyle Pr ( \bm{\kappa} \in \mathcal{K}  | \rfpe{\mathcal{M}, \mathcal{D}^{(l)}}{*} )\)},
width=\figW,
x grid style={white!69.0196078431373!black},
xlabel={data acquisition step \(\displaystyle l\)},
xmin=-0.2, xmax=4.2,
xtick style={color=black},
y grid style={white!69.0196078431373!black},
ylabel={\(\displaystyle Pr \left( \bm{\kappa} \in \mathcal{K}  \middle| \varphi^*_{\mathcal{M}, \mathcal{D}^{(l)}} \right)\)  },
ymin=0.3125, ymax=0.65625,
ytick style={color=black}
]
\addplot [thick, color0, mark=*, mark size=1.5, mark options={solid}]
table {%
0 0.328125
1 0.4765625
2 0.607421875
3 0.634765625
4 0.640625
};
\end{axis}

\end{tikzpicture}
\subcaption{Convergence of $Pr ( \hp \in \mathcal{K} | \varphi^*_{\mathcal{M}, \mathcal{D}^{(l)}} )$} \label{fig:active_learning_convergence}
\end{minipage}%
\caption{\textbf{Case 1: Convergence characteristics  of the optimization algorithm.} \textit{(a)} evolution of the ELBO $\mathcal{F}$ as a function of the iteration number in the inner loop (see Algorithm \ref{Alg:EM}) and for $l=0$ (outer loop - see Algorithm \ref{Alg:EM}). \textit{(b)} evolution of the probability we seek to maximize $Pr\left[ \hp \in \mathcal{K} | \varphi \right]$ (estimated with $1024$ Monte Carlo samples and the high-fidelity model) at the optimal $\rfp$ values identified by the algorithm at various data acquisition steps $l$ (outer loop - see Algorithm \ref{Alg:EM}).
}
\label{fig:inner-outer-convergence}
\end{figure*}
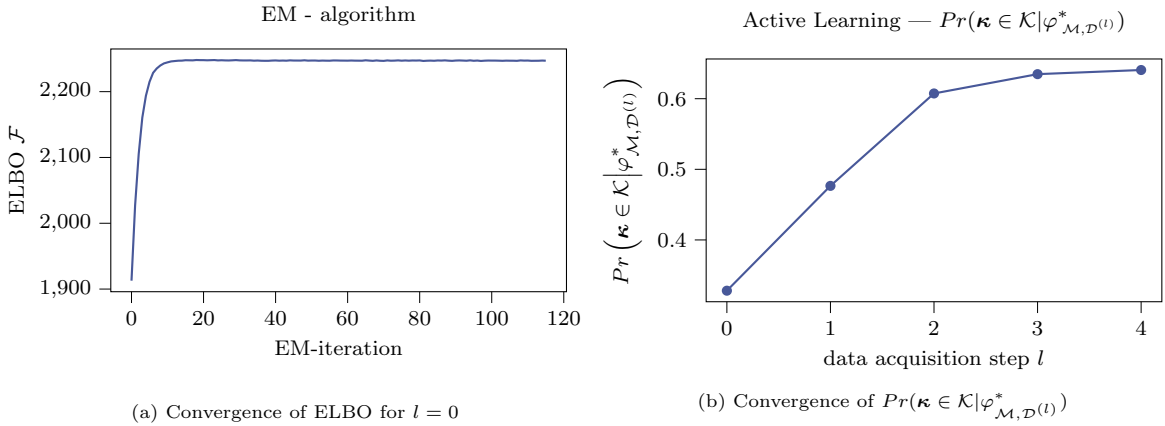

The optimal process parameters at each data acquisition step are denoted as $\rfpe{\mathcal{M}, \mathcal{D}^{(l)}}{*}$, with the subscript indicating the dependence on the surrogate model $\mathcal{M}$ and the dataset $\mathcal{D}^{(l)}$ on which it has been trained. Once the algorithm has converged to its final estimate of the process parameters after $L$ data acquisition steps, i.e. $\rfpe{\mathcal{M}, \mathcal{D}^{(L)}}{*}$, we can assess $\rfpe{\mathcal{M}, \mathcal{D}^{(L)}}{*}$ by obtaining a \emph{reference} estimate of the expected utility $U ( \rfpe{\mathcal{M}, \mathcal{D}^{(L)}}{*} )  = Pr ( \hp \in \mathcal{K} | \rfpe{\mathcal{M}, \mathcal{D}^{(L)}}{*} )$ using a Monte Carlo simulation, i.e. by sampling microstructures $\bm{x} \sim p ( \bm{x} | \rfpe{\mathcal{M}, \mathcal{D}^{(L)}}{*} )$,  and running the high-fidelity model instead of the inexpensive surrogate.  
This approach will also enable us to compare the optimization results obtained with the active learning strategy 
 with those obtained by using randomized training data $\mathcal{D}$ for the surrogate (i.e. without adaptive learning). 
We argue that the former has a competitive advantage, if for the same number of datapoints $N$ we can achieve a higher score in terms of our materials design objective $Pr \left( \hp \in \mathcal{K} \middle| \varphi^* \right)$. As the optimization objective $\mathcal{F}$ is non-convex and the optimization algorithm itself non-deterministic \footnote{due to the randomized generation of the data, the stochastic initialization of the neural network, the randomized initial guess of $\varphi^{(0)} \sim \mathcal{N} \left( \bm{0}, \mat{I} \right)$}, generally the process parameters $\varphi^*$ identified can vary across different runs. For this reason the optimization problem is solved several times (with different randomized initializations) and we report on the aggregate performance of active learning vs. randomized data generation (serving as a baseline). \\

In the following we discuss the results obtained and displayed in Fig. \ref{fig:multiphysics_microstructures},  \ref{fig:multiphysics_distribution2d},  \ref{fig:MultiphysicsHybridPlot} and \ref{fig:inner-outer-convergence}.

\begin{itemize}
\item In Fig. \ref{fig:multiphysics_microstructures} we depict sample  microstructures  drawn from $p(\bx|\rfp)$ for two values of $\rfp$ i.e. for the initial guess $\varphi^{(0)}$ (top row) and for  optimal process parameters $\rfpe{\mathcal{M}, \mathcal{D}^{(L)}}{*}$ (bottom row). 
While the microstructures in the bottom row  remain random, one observes that  the connectivity of phase 1 (stiffer) is increased as compared to the microstructures of the top row. The diagonal, connected paths of the lesser conducting phase (black) effectively block heat conduction in the horizontal direction. 
This is reflected in the effective properties reported underneath each image. The value of the objective, i.e. of the probability that the properties $\hp \in \mathcal{K}$, is $\approx 0.65$ for the microstructures in the bottom row (see Fig. \ref{fig:MultiphysicsHybridPlot} - right)  whereas for the top ones $\approx 0.14$. 

\item Fig. \ref{fig:multiphysics_distribution2d} provides insight into the optimization algorithm proposed by looking at the process-property density $p(\hp |\rfp)$ for various $\rfp$ values. We note that this is implicitly defined by propagating the randomness in the microstructures (quantified by $p(\bx|\rfp)$) through the high-fidelity model that predicts the properties of interest. Based on the  Monte Carlo estimates depicted in Fig. \ref{fig:multiphysics_distribution2d}, one observes that the density $p(\bm{\kappa}| \rfp)$ which only minimally touches upon the target domain $\mathcal{K}$ for initial process parameters $\varphi^{(0)}$ (left), gradually moves closer to $\mathcal{K}$ as the iterations proceed and by using the surrogate trained on the initial batch of data $\mathcal{D}^{(0)}$ (middle). The incorporation of additional training data through the successive applications of the adaptive learning scheme described earlier, enables the surrogate  to sufficiently resolve details in the structure-property map that eventually lead to the density $p(\hp | \rfp)$ shown on the right panel and which maximally overlaps (in comparison) with the target domain $\mathcal{K}$. 
\item On the left side of Fig. \ref{fig:MultiphysicsHybridPlot} we illustrate the performance advantage gained by the active learning approach proposed over a brute-force strategy that employs randomized training data for the surrogate.  To this end, we compare the values of the objective function, i.e.  $Pr ( \hp \in \mathcal{K} | \rfpe{\mathcal{M}, \mathcal{D}}{*} )$ achieved for datasets $\mathcal{D}$ of equal size, with the dataset being either generated randomly, or constructed based on our active learning approach. Evidently, the active learning approach was able to achieve a better material design at comparably significantly lower numerical cost (as measured by the number of evaluations of the high-fidelity model of the microstructure-property link). We observe that while the addition of more training data generally leads to more accurate surrogates, when this is done without regard to the optimization objectives (red line) then it does not necessarily lead to higher values of the objective function.
On the right panel of Fig. \ref{fig:MultiphysicsHybridPlot} we  provide further insight as to why the adaptive data acquisition was able to outperform a randomized approach. To this end and for one of the two properties, we compare the model-based belief $p \left( \kappa_1 \middle| \rfpe{\mathcal{M}, \mathcal{D}^{(4)}}{*}, \mathcal{D}^{(0)} \right)$ of the surrogate conditional on $\mathcal{D}^{(0)}$ against a reference density obtained using Monte Carlo (black line). We can see that a model only informed by $\mathcal{D}^{(0)}$ (blue line) identifies an incorrect  density  and as such fails to converge to the optimal process parameters. The active learning approach (red line) was able to correct the initially erroneous model belief and as a result performs better in the optimization task.
\item In Fig. \ref{fig:elbo_convergence} we illustrate the  evolution of the ELBO during the inner-loop iterations of the proposed VB-EM algorithm (see Algorithm \ref{Alg:EM}).
 Finally, in Fig. \ref{fig:active_learning_convergence} we depict (on the right) the evolution of the maximum of the objective identified at various data acquisition steps $l$  of the proposed  active learning scheme. As it can be seen, the targeted data enrichment enables the surrogate to resolve  details in the structure-property map  and identify higher-performing processing parameters $\rfp$.
\end{itemize}


\subsubsection*{Case 2: Target density of properties (O2)}

\begin{figure*}[hbtp]
    \centering

       \includegraphics[height=6.1cm, width=0.975\linewidth]{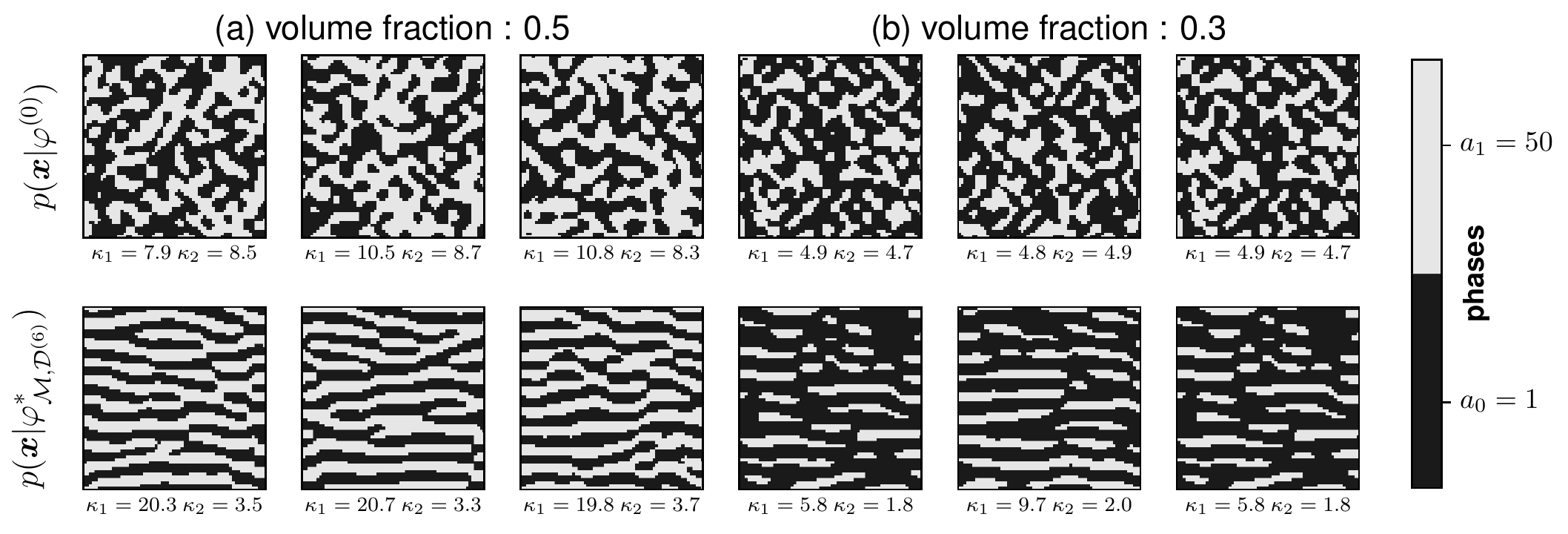}
    \caption{ 
    \textbf{Case 2: Optimal random microstructures.} \textit{(Tow row)} Samples of microstructures drawn from $p(\bx|\rfp)$ for the initial guess $\varphi^{(0)}$ of processing variables, \textit{(Bottom row)} Samples of microstructures drawn from $p(\bx|\rfp)$ for the optimal value $\rfpe{\mathcal{M}, \mathcal{D}^{(6)}}{*}$ of the  processing variables  which minimize the KL-divergence between  $p(\hp|\rfp)$  and the target density $p_{target}$ (\refeqp{Eq:GaussianTargetDistribution}). Underneath each microstructure, the thermal properties $\kappa_{1}, \kappa_{2}$   of interest (\refeqp{eq:prop2}) are reported.
    The illustrations correspond to two volume fractions $0.5$ (in \textit{(a)}-left) and $0.3$ (in \textit{(b)}-right) of the high-conductivity phase ($a_1 = 50$).
    }
    \label{fig:channelizedflow64_microstructure_evolution}
    \vspace{0.15cm}
    
           \includegraphics[width=0.92\linewidth]{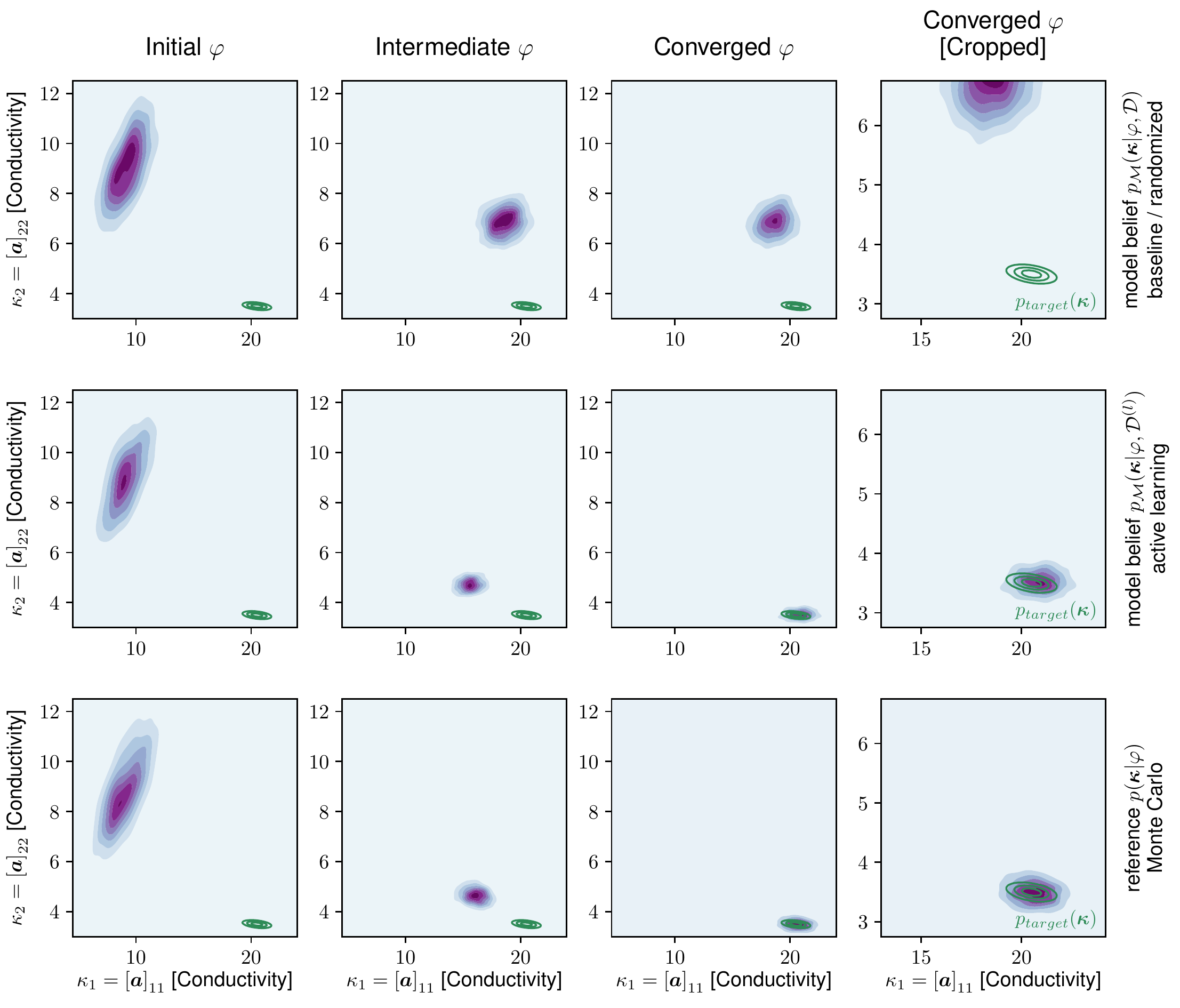}
   \caption{\textbf{Case 2: Evolution the  of process-property  density    $p(\hp | \rfp)$ with and without active learning in relation to the target  $p_{target} \left( \hp \right)$. }
    We plot the evolution of the process-property  density $p(\hp | \rfp)$ at three  different stages of each optimization run, i.e. for the initial $\rfp$ (left column) for the $\rfp$ at an intermediate stage of the optimization (middle column) and for  the optimal $\rfp$ identified upon convergence (right column). The fourth column is a zoomed-in version of the third that enables closer comparisons of the densities involved.
     \textit{(Top row:)} illustrates $p(\hp | \rfp)$ as predicted by the surrogate trained on a randomized dataset without active learning.  \textit{(Middle row:)} illustrates $p(\hp | \rfp)$ as predicted by the surrogate trained using the adaptive learning proposed. \textit{Bottom row:} illustrates  the actual $p(\hp | \rfp)$ (estimated with $1024$ Monte Carlo samples and the high-fidelity model) and for the optimal $\rfp$  identified by the active learning approach. The target distribution $p_{target} \left( \hp \right)$ is indicated with green iso-probability lines.
     }
    \label{fig:channelizedflow64_jointpdf_evolution}
\end{figure*}

In this second numerical illustration, we investigate the performance of the proposed methodological framework  for an (O2)-type optimization problem (\refeqp{eq:o2}) where we seek to {\em identify the processing parameters $\rfp$ that lead to a property density $p(\hp | \rfp)$ that is closest to a prescribed target $p_{target} \left( \hp \right)$}. In particular, we considered the following two properties

\begin{align}
    \kappa_1 = \left[ \bm{a}^{\text{eff}} \right]_{11} \qquad \qquad \qquad \kappa_2 = \left[ \bm{a}^{\text{eff}} \right]_{22}
    \label{eq:prop2}
\end{align}
i.e. $\hp \in \mathbb{R}^2$ and a target density:
\begin{align}
p_{target} \left( \hp \right) = \mathcal{N} \left( \hat{\bm{\mu}}, \mat{\hat{\Sigma}} \right) 
\label{Eq:GaussianTargetDistribution}
\end{align}
with $\bm{\hat{\mu}} = \left[ 20.5, 3.5 \right]^T$ and $\hat{\Sigma}_{11} = 0.60$, $\hat{\Sigma}_{22} = 0.01, \hat{\Sigma}_{12} = -0.03$ (indicated with green iso-probability lines in Fig. \ref{fig:channelizedflow64_jointpdf_evolution}). These values were selected to promote anisotropic behavior, i.e. microstructures are desired to have a large effective conductivity in the first spatial dimension, while  being comparatively insulating in the second spatial dimension.
The characteristics of the active learning procedure (outer loop in Algorithm \ref{Alg:EM}) remain identical, with the only difference that $\mathcal{D}^{(0)}$ now comprises \tmp{$N_0 = 4096$} datapoints, with \tmp{$N_{add} = 1024$} datapoints (out of \tmp{$4096$} candidates) added in each of the \tmp{$L=6$} data-enrichment steps. We used $S=20$ samples from $p_{target} \left( \hp \right)$ to approximate the objective (see Eq. \eqref{Eq:ELBO_O2}).

We present and discuss the results obtained based on the Figures \ref{fig:channelizedflow64_microstructure_evolution} and \ref{fig:channelizedflow64_jointpdf_evolution}:

\begin{itemize}

\item In Fig. \ref{fig:channelizedflow64_microstructure_evolution} we showcase sample  microstructures drawn from $p \left( \bm{x} \middle| \varphi \right)$ both for the initial guess $\varphi^{(0)}$ (top row) as well as for the optimal process parameters $\rfpe{\mathcal{M}, \mathcal{D}^{(L)}}{*}$ identified by the optimized algorithm using the active learning approach. The leftmost three columns pertain to volume fraction $0.5$ (of the more conducting phase $a_1$) whereas the rightmost three columns to volume fraction $0.3$.

As one would expect, we observe that the optimal family of microstructures identified (determined by $\rfpe{\mathcal{M}, \mathcal{D}^{(L)}}{*}$) 
exhibit connected paths of the more conductive phase (white) along the horizontal direction. The connected paths of the lesser conducting phase (black) are also aligned in the horizontal direction so as to reduce the effective conductivity along the vertical direction. 
The optimal microstructures therefore exhibit a strong anistropic property by funneling heat through pipe-like structures of high-conductivity material in the horizontal direction. This is also reflected in the indicative property values report under each image.

\item Finally,  Fig. \ref{fig:channelizedflow64_jointpdf_evolution} assesses the advantage of the active learning strategy advocated for this problem. 
In particular, we plot the evolution of the process-structure density $p \left( \bm{\kappa} \middle| \varphi \right)$ in relation to the target  $p_{target} \left( \hp \right)$ (depicted with \tmp{green} iso-probability lines) at different stages of each optimization (initial-intermediate-converged).
Using the optimal process parameters $\varphi$ at each of these stages, we see that the optimization scheme {\em without} active learning (top row) results in a density that is quite far from the target.
In contrast, the optimization algorithm {\em with} active learning (middle row) is able to identify a $\rfp$  which brings the  $p \left( \bm{\kappa} \middle| \varphi \right)$  into close alignment with the target distribution $p_{target} \left( \hp \right)$.
The validity of this result is assessed on the bottom row where the actual $p(\hp| \rfp)$ (estimated with Monte Carlo and the high-fidelity model) is depicted for the $\rfp$ values identified by the active learning approach in the middle row. We observe a very close agreement which reinforces the evidence that the active learning strategy advocated enables the surrogate to accurately resolve the details in the structure-property map that are needed for the solution of the optimization problem.
\end{itemize}

In summary, we presented  a flexible, fully probabilistic, data-driven  formulation for materials design that accounts  for the uncertainty in the process-structure and structure-property links and enables the identification  of the optimal, high-dimensional, process parameters $\varphi$.
We have demonstrated that a variety of different objectives can be accommodated.
The methodology relies on the availability of a process-structure linkage which we generically represented with the density $p(\bx|\rfp)$ and could be learned from experimental or simulation data. 
It is not restricted to the particular parametrizations adopted in terms of the process $\rfp$ or microstructural $\bx$ variables and very high-dimensional descriptions can be employed due to the VB-EM scheme advocated.
We have also demonstrated the use of probabilistic surrogates in combination with novel, active learning formulations which can significantly reduce the computational effort associated with the  structure-property link and enable the solution of problems with a small number of such simulations.
The optimization framework is also agnostic regarding the nature of the structure-property link $p \left( \hp \middle| \bm{x} \right)$ (and whether it is deterministic or probabilistic), as it simply defines the data generation process for the probabilistic surrogate. As  a result, other material properties or other physical descriptions can be readily incorporated.
The framework advocated does not rely on a particular architecture of the data-driven surrogate. Its predictive uncertainty is incorporated and the self-supervised active learning mechanism can control the number of training data that each particular surrogate would need. 
While not discussed, it is also possible to assess the optimization error, albeit with additional runs of the high-fidelity model, by using an Importance Sampling step \cite{sternfels2011stochastic}.
Lastly we mention further potential of improvement by a fully Bayesian treatment of the  surrogate's parameters $\bm{\theta}$, which would be particularly beneficial in the small-data regime we are operating in.

\begin{figure}[t]
\begin{algorithm}[H]
\setstretch{1.25}

  \caption{Obtain $\varphi^* = \arg \max_{\varphi} U_{1,2}$ (Eq. (\ref{eq:obj1}) or Eq. (\ref{eq:o2}))
    using a probabilistic surrogate and active learning}
 \label{Alg:EM}
  \KwData{$l=0$, $t=0$ $\mathcal{D}^{(0)}$, structure-property-model, surrogate $p \left( \hp \middle| \bm{x}, \mathcal{D}^{(0)} \right)$, initial $\varphi^{(0)}$, variational family $\mathcal{Q}_{\bm{\xi}}$}  
  \KwResult{Converged process parameter $\varphi^*_{\mathcal{D}^{(L)}}$}  
 \While{$\varphi$ not converged}{
 \While{ELBO not converged}{
 
   \tcc{Execute E-step}
 \begin{equation*}
  \bm{\xi}^{(t+1)} = \arg \max_{\bm{\xi}} \mathcal{F} \left( \varphi^{(t)}, q_{\bm{\xi}} \left( \hp, \bm{x} \right)  \right)
\end{equation*}
  
  \tcc{Execute M-step}
  
   \begin{equation*}
    \varphi^{(t+1)} =\arg \max_{\varphi} \mathcal{F} \left( \varphi, q_{\bm{\xi}^{(t+1)}} \left( \hp, \bm{x} \right) \right) 
\end{equation*}
  
$t \to t+1$ \;
 }
\tcc{Optimal $\varphi$ conditional on current data}
\vspace{0.05cm}
$~~~~~ \varphi^*_{\mathcal{M}, \mathcal{D}^{(l)}} \leftarrow \varphi^{(t)} $

\vspace{0.2cm}
\tcc{Create microstructure candidates}
sample $\bm{x}^{(l,n)} \sim q \left( \bm{x} \right), ~~ n=1, \ldots, N_{pool}$\;

compute $\alpha(\bm{x}^{(l,n)})$ (Eq. \eqref{eq:acquisition})\;

\vspace{0.2cm}
 \tcc{Select most informative subset}
Select $N_{add}<N_{pool}$ microstructures from $\mathcal{D}_{pool}^{(l)}$ which yield the highest acquisition function values and compute the corresponding property values $\hp(\bx)$ in order to form $\mathcal{D}_{add}^{(l)}$.\;
 
 
 \begin{align*}
     \mathcal{D}^{(l+1)} \leftarrow \mathcal{D}^{(l)} \cup \mathcal{D}_{add}^{(l)}
 \end{align*}
 
  \tcc{Update probabilistic surrogate}
 
 \begin{align*}
     p \left( \bm{\kappa} \middle| \bm{x} , \mathcal{D}^{(l+1)} \right) \leftarrow  p \left( \bm{\kappa} \middle| \bm{x} , \mathcal{D}^{(l)} \right)  
 \end{align*}
 
 $l \to l+1$

 }
\end{algorithm}
\caption{\textbf{Pseudo-code for proposed algorithm.} The inner VB-EM iterations are wrapped within the adaptive data acquisition as an outer loop.}
\end{figure}

\section*{Code Availability}

The source code will be made available at \url{https://github.com/bdevl/SMO}.

\section*{Data Availability}

The accompanying data will be made available at \url{https://github.com/bdevl/SMO}.

\section*{Author Contributions}

M.R.: conceptualization, physics and machine-learning modeling and computations, algorithmic and code development, writing of the paper. P-S.K: conceptualization, writing of the paper

\section*{Competing Interests}

The authors declare no competing interests.

\bibliographystyle{unsrt}
\bibliography{references}

\begin{thebibliography}{10}

\bibitem{mgi_2011}
National Science {and} Technology~Council Executive Office of~the President.
\newblock {\em Materials {Genome} {Initiative}: {A} {Renaissance} of {American}
  {Manufacturing}}.
\newblock June 2011.

\bibitem{mcdowell_integrated_2009}
David~L McDowell, Jitesh Panchal, Hae-Jin Choi, Carolyn Seepersad, Janet Allen,
  and Farrokh Mistree.
\newblock {\em Integrated design of multiscale, multifunctional materials and
  products}.
\newblock Butterworth-Heinemann, 2009.

\bibitem{arroyave_systems_2019}
Raymundo Arróyave and David~L. McDowell.
\newblock Systems {Approaches} to {Materials} {Design}: {Past}, {Present}, and
  {Future}.
\newblock {\em Annual Review of Materials Research}, 49(1):103--126, 2019.
\newblock \_eprint: https://doi.org/10.1146/annurev-matsci-070218-125955.

\bibitem{chernatynskiy_uncertainty_2013}
Aleksandr Chernatynskiy, Simon~R. Phillpot, and Richard LeSar.
\newblock Uncertainty {Quantification} in {Multiscale} {Simulation} of
  {Materials}: {A} {Prospective}.
\newblock {\em Annual Review of Materials Research}, 43(1):157--182, 2013.

\bibitem{honarmandi_uncertainty_2020}
Pejman Honarmandi and Raymundo Arróyave.
\newblock Uncertainty {Quantification} and {Propagation} in {Computational}
  {Materials} {Science} and {Simulation}-{Assisted} {Materials} {Design}.
\newblock {\em Integrating Materials and Manufacturing Innovation},
  9(1):103--143, March 2020.

\bibitem{liu_xuan_nasa_2018}
{Liu, Xuan}, {Furrer, David}, {Kosters, Jared}, and {Holmes, Jack}.
\newblock {NASA} {Vision} 2040: {A} {Roadmap} for {Integrated}, {Multiscale}
  {Modeling} and {Simulation} of {Materials} and {Systems}.
\newblock Technical report, March 2018.

\bibitem{bock_review_2019}
Frederic~E. Bock, Roland~C. Aydin, Christian~J. Cyron, Norbert Huber, Surya~R.
  Kalidindi, and Benjamin Klusemann.
\newblock A {Review} of the {Application} of {Machine} {Learning} and {Data}
  {Mining} {Approaches} in {Continuum} {Materials} {Mechanics}.
\newblock {\em Frontiers in Materials}, 6, 2019.
\newblock Publisher: Frontiers.

\bibitem{panchal_key_2013}
Jitesh~H. Panchal, Surya~R. Kalidindi, and David~L. McDowell.
\newblock Key computational modeling issues in {Integrated} {Computational}
  {Materials} {Engineering}.
\newblock {\em Computer-Aided Design}, 45(1):4--25, January 2013.

\bibitem{grigo_bayesian_2019}
Constantin Grigo and Phaedon-Stelios Koutsourelakis.
\newblock Bayesian {Model} and {Dimension} {Reduction} for {Uncertainty}
  {Propagation}: {Applications} in {Random} {Media}.
\newblock {\em SIAM/ASA Journal on Uncertainty Quantification}, 7(1):292--323,
  January 2019.

\bibitem{zabaras_scalable_2008}
N.~Zabaras and B.~Ganapathysubramanian.
\newblock A scalable framework for the solution of stochastic inverse problems
  using a sparse grid collocation approach.
\newblock {\em Journal of Computational Physics}, 227(9):4697--4735, April
  2008.

\bibitem{frazier_bayesian_2016}
Peter~I Frazier and Jialei Wang.
\newblock Bayesian optimization for materials design.
\newblock In {\em Information {Science} for {Materials} {Discovery} and
  {Design}}, pages 45--75. Springer, 2016.

\bibitem{zhang_bayesian_2020}
Yichi Zhang, Daniel~W. Apley, and Wei Chen.
\newblock Bayesian {Optimization} for {Materials} {Design} with {Mixed}
  {Quantitative} and {Qualitative} {Variables}.
\newblock {\em Scientific Reports}, 10(1):4924, March 2020.
\newblock Number: 1 Publisher: Nature Publishing Group.

\bibitem{jung_microstructure_2020}
Jaimyun Jung, Jae~Ik Yoon, Hyung~Keun Park, Hyeontae Jo, and Hyoung~Seop Kim.
\newblock Microstructure design using machine learning generated low
  dimensional and continuous design space.
\newblock {\em Materialia}, 11:100690, June 2020.

\bibitem{chen_machine_2019}
Chun-Teh Chen and Grace~X. Gu.
\newblock Machine learning for composite materials.
\newblock {\em MRS Communications}, 9(2):556--566, June 2019.
\newblock Publisher: Cambridge University Press.

\bibitem{torquato_optimal_2010}
S.~Torquato.
\newblock Optimal {Design} of {Heterogeneous} {Materials}.
\newblock {\em Annual Review of Materials Research}, 40(1):101--129, 2010.

\bibitem{hoffman_stochastic_2013}
Matthew~D. Hoffman, David~M. Blei, Chong Wang, and John Paisley.
\newblock Stochastic {Variational} {Inference}.
\newblock {\em J. Mach. Learn. Res.}, 14(1):1303--1347, May 2013.

\bibitem{tran_solving_2021}
Anh Tran and Tim Wildey.
\newblock Solving stochastic inverse problems for property--structure linkages
  using data-consistent inversion and machine learning.
\newblock {\em JOM}, 73(1):72--89, January 2021.

\bibitem{nosouhi_dehnavi_framework_2020}
Fayyaz Nosouhi~Dehnavi, Masoud Safdari, Karen Abrinia, Ali Hasanabadi, and
  Majid Baniassadi.
\newblock A framework for optimal microstructural design of random
  heterogeneous materials.
\newblock {\em Computational Mechanics}, 66(1):123--139, July 2020.

\bibitem{acar_stochastic_2017}
Pınar Acar, Siddhartha Srivastava, and Veera Sundararaghavan.
\newblock Stochastic {Design} {Optimization} of {Microstructures} with
  {Utilization} of a {Linear} {Solver}.
\newblock {\em AIAA Journal}, 55(9):3161--3168, 2017.
\newblock Publisher: American Institute of Aeronautics and Astronautics
  \_eprint: https://doi.org/10.2514/1.J056000.

\bibitem{acar_stochastic_2019}
Pinar Acar and Veera Sundararaghavan.
\newblock Stochastic {Design} {Optimization} of {Microstructural} {Features}
  {Using} {Linear} {Programming} for {Robust} {Design}.
\newblock {\em AIAA Journal}, 57(1):448--455, 2019.
\newblock Publisher: American Institute of Aeronautics and Astronautics
  \_eprint: https://doi.org/10.2514/1.J057377.

\bibitem{liu_predictive_2015}
Ruoqian Liu, Abhishek Kumar, Zhengzhang Chen, Ankit Agrawal, Veera
  Sundararaghavan, and Alok Choudhary.
\newblock A predictive machine learning approach for microstructure
  optimization and materials design.
\newblock {\em Scientific reports}, 5(1):1--12, 2015.
\newblock Publisher: Nature Publishing Group.

\bibitem{paul_microstructure_2019}
Arindam Paul, Pinar Acar, Wei-keng Liao, Alok Choudhary, Veera Sundararaghavan,
  and Ankit Agrawal.
\newblock Microstructure optimization with constrained design objectives using
  machine learning-based feedback-aware data-generation.
\newblock {\em Computational Materials Science}, 160:334--351, April 2019.

\bibitem{popova_process-structure_2017}
Evdokia Popova, Theron~M. Rodgers, Xinyi Gong, Ahmet Cecen, Jonathan~D.
  Madison, and Surya~R. Kalidindi.
\newblock Process-{Structure} {Linkages} {Using} a {Data} {Science} {Approach}:
  {Application} to {Simulated} {Additive} {Manufacturing} {Data}.
\newblock {\em Integrating Materials and Manufacturing Innovation},
  6(1):54--68, March 2017.

\bibitem{lee_fast_2021}
Xian~Yeow Lee, Joshua~R. Waite, Chih-Hsuan Yang, Balaji Sesha~Sarath Pokuri,
  Ameya Joshi, Aditya Balu, Chinmay Hegde, Baskar Ganapathysubramanian, and
  Soumik Sarkar.
\newblock Fast inverse design of microstructures via generative invariance
  networks.
\newblock {\em Nature Computational Science}, 1(3), March 2021.

\bibitem{ikebata_bayesian_2017}
Hisaki Ikebata, Kenta Hongo, Tetsu Isomura, Ryo Maezono, and Ryo Yoshida.
\newblock Bayesian molecular design with a chemical language model.
\newblock {\em Journal of Computer-Aided Molecular Design}, 31(4):379--391,
  April 2017.

\bibitem{dempster_maximum_1977}
A.~P. Dempster, N.~M. Laird, and D.~B. Rubin.
\newblock Maximum likelihood from incomplete data via the {EM} algorithm (with
  discussion).
\newblock {\em Journal of the Royal Statistical Society B}, 39(1):1--38, 1977.

\bibitem{beal_variational_2003}
Matthew~J Beal and Zoubin Ghahramani.
\newblock The {Variational} {Bayesian} {EM} {Algorithm} for {Incomplete}
  {Data}: with {Application} to {Scoring} {Graphical} {Model} {Structures}.
\newblock {\em Bayesian Statistics}, (7), 2003.

\bibitem{neal1998view}
Radford~M Neal and Geoffrey~E Hinton.
\newblock A view of the {EM} algorithm that justifies incremental, sparse, and
  other variants.
\newblock In {\em Learning in graphical models}, pages 355--368. Springer,
  1998.

\bibitem{kalidindi_bayesian_2019}
Surya~R. Kalidindi.
\newblock A {Bayesian} framework for materials knowledge systems.
\newblock {\em MRS Communications}, 9(2):518--531, June 2019.
\newblock Publisher: Cambridge University Press.

\bibitem{marcus_rebooting_2019}
Gary Marcus and Ernest Davis.
\newblock {\em Rebooting {AI}: {Building} {Artificial} {Intelligence} {We}
  {Can} {Trust}}.
\newblock Pantheon, September 2019.

\bibitem{yang2018deep}
Zijiang Yang, Yuksel~C Yabansu, Reda Al-Bahrani, Wei-keng Liao, Alok~N
  Choudhary, Surya~R Kalidindi, and Ankit Agrawal.
\newblock Deep learning approaches for mining structure-property linkages in
  high contrast composites from simulation datasets.
\newblock {\em Computational Materials Science}, 151:278--287, 2018.

\bibitem{cecen2018material}
Ahmet Cecen, Hanjun Dai, Yuksel~C Yabansu, Surya~R Kalidindi, and Le~Song.
\newblock Material structure-property linkages using three-dimensional
  convolutional neural networks.
\newblock {\em Acta Materialia}, 146:76--84, 2018.

\bibitem{tong_active_2001}
Simon Tong and Stanford University Computer~Science Dept.
\newblock {\em Active learning: theory and applications}.
\newblock Stanford University, 2001.

\bibitem{mackay_information-based_1992}
D.~J.~C. MacKay.
\newblock Information-{Based} {Objective} {Functions} for {Active} {Data}
  {Selection}.
\newblock {\em Neural Computation}, 4(4):590--604, July 1992.
\newblock Conference Name: Neural Computation.

\bibitem{sahoo2017online}
Doyen Sahoo, Quang Pham, Jing Lu, and Steven~CH Hoi.
\newblock Online deep learning: Learning deep neural networks on the fly.
\newblock {\em arXiv preprint arXiv:1711.03705}, 2017.

\bibitem{teubner_level_1991}
M~Teubner.
\newblock Level surfaces of {Gaussian} random fields and microemulsions.
\newblock {\em EPL (Europhysics Letters)}, 14(5):403, 1991.
\newblock Publisher: IOP Publishing.

\bibitem{roberts_transport_1995}
Anthony~P Roberts and Max Teubner.
\newblock Transport properties of heterogeneous materials derived from
  {Gaussian} random fields: bounds and simulation.
\newblock {\em Physical Review E}, 51(5):4141, 1995.
\newblock Publisher: APS.

\bibitem{koutsourelakis_probabilistic_2006}
P.S. Koutsourelakis.
\newblock Probabilistic characterization and simulation of multi-phase random
  media.
\newblock {\em Probabilistic Engineering Mechanics}, 21(3), 2006.

\bibitem{bostanabad_stochastic_2016}
Ramin Bostanabad, Anh~Tuan Bui, Wei Xie, Daniel~W. Apley, and Wei Chen.
\newblock Stochastic microstructure characterization and reconstruction via
  supervised learning.
\newblock {\em Acta Materialia}, 103:89--102, January 2016.

\bibitem{cang_microstructure_2017}
Ruijin Cang, Yaopengxiao Xu, Shaohua Chen, Yongming Liu, Yang Jiao, and Max
  Yi~Ren.
\newblock Microstructure {Representation} and {Reconstruction} of
  {Heterogeneous} {Materials} {Via} {Deep} {Belief} {Network} for
  {Computational} {Material} {Design}.
\newblock {\em Journal of Mechanical Design}, 139(7), May 2017.

\bibitem{miehe2002computational}
Christian Miehe and Andreas Koch.
\newblock Computational micro-to-macro transitions of discretized
  microstructures undergoing small strains.
\newblock {\em Archive of Applied Mechanics}, 72(4):300--317, 2002.

\bibitem{hill1972constitutive}
Rodney Hill.
\newblock On constitutive macro-variables for heterogeneous solids at finite
  strain.
\newblock {\em Proceedings of the Royal Society of London. A. Mathematical and
  Physical Sciences}, 326(1565):131--147, 1972.

\bibitem{saheli_microstructure_2004}
G.~Saheli, H.~Garmestani, and B.~L. Adams.
\newblock Microstructure design of a two phase composite using two-point
  correlation functions.
\newblock {\em Journal of Computer-Aided Materials Design}, 11(2):103--115,
  January 2004.

\bibitem{fullwood_microstructure_2010}
David~T. Fullwood, Stephen~R. Niezgoda, Brent~L. Adams, and Surya~R. Kalidindi.
\newblock Microstructure sensitive design for performance optimization.
\newblock {\em Progress in Materials Science}, 55(6):477--562, August 2010.

\bibitem{sternfels2011stochastic}
Raphael Sternfels and Phaedon-Stelios Koutsourelakis.
\newblock Stochastic design and control in random heterogeneous materials.
\newblock {\em International Journal for Multiscale Computational Engineering},
  9(4), 2011.

\bibitem{wilson2013gaussian}
Andrew Wilson and Ryan Adams.
\newblock Gaussian process kernels for pattern discovery and extrapolation.
\newblock In {\em International conference on machine learning}, pages
  1067--1075. PMLR, 2013.

\bibitem{shinozuka1996simulation}
Masanobu Shinozuka and George Deodatis.
\newblock Simulation of multi-dimensional gaussian stochastic fields by
  spectral representation.
\newblock 1996.

\bibitem{hu1997simulation}
B~Hu and W~Schiehlen.
\newblock On the simulation of stochastic processes by spectral representation.
\newblock {\em Probabilistic engineering mechanics}, 12(2):105--113, 1997.

\bibitem{kingma2013auto}
Diederik~P Kingma and Max Welling.
\newblock Auto-encoding variational bayes.
\newblock {\em arXiv preprint arXiv:1312.6114}, 2013.

\bibitem{kingma2014adam}
Diederik~P Kingma and Jimmy Ba.
\newblock Adam: A method for stochastic optimization.
\newblock {\em arXiv preprint arXiv:1412.6980}, 2014.

\bibitem{paszke2019pytorch}
Adam Paszke, Sam Gross, Francisco Massa, Adam Lerer, James Bradbury, Gregory
  Chanan, Trevor Killeen, Zeming Lin, Natalia Gimelshein, Luca Antiga, et~al.
\newblock Pytorch: An imperative style, high-performance deep learning library.
\newblock {\em Advances in neural information processing systems},
  32:8026--8037, 2019.

\bibitem{smith2013sequential}
Adrian Smith.
\newblock {\em Sequential Monte Carlo methods in practice}.
\newblock Springer Science \& Business Media, 2013.

\end{thebibliography}

\clearpage
\section*{Supplementary Information}

In the following, we provide details specific to the algorithm implementation and the numerical simulations.

\subsection*{Process-Structure linkage}
As mentioned earlier the discretized, two-phase random microstructures  employed in the numerical illustrations are  represented by a random vector $\bx$ which arises by thresholding a two-dimensional, zero-mean, unit-variance  Gaussian field, in its discretized form denoted by the vector $\bx_g$. The cutoff threshold $x_0$ is specified by the desired volume fraction and the parameters $\rfp$ are associated with the   spectral density function (SDF) $G \left( \bm{w} \right)$ of the underlying Gaussian field. The SDF $G \left( \bm{w} \right)$ arises as the Fourier dual of the autocovariance, where $\bs{w}=[w_1,w_2]^T\in \RR^2$ denotes the wavenumbers.
We express the SDF as:
\begin{align}
G \left( \vect{w} \right) = \sum\limits_{i=1}^Q \gamma_i h_i \left( \vect{w} ; \vect{\mu}_i, \sigma_i \right)
    \label{Eq:spectral_mixture_kernel}
\end{align}
where the functions $h_i$ are Radial Basis Functions (RBFs) which depend on the parameters $ \vect{\mu}_i, \sigma_i $ and have the functional form:
\begin{align}
h\left( \vect{w} ; \vect{\mu}, \sigma \right)=\frac{1}{\sqrt{2\pi \sigma^2}}e^{-\frac{1}{2\sigma^2} ||\bs{w}-\bs{\mu}||^2}.
\end{align}
This form is adopted because it automatically ensures the positivity of the resulting SDF. Eq. \eqref{Eq:spectral_mixture_kernel} is also known as a \emph{spectral mixture kernel}, which defines a universal approximator for sufficiently large $Q$  \cite{wilson2013gaussian}.
In our simulations,  the parameters $\{\vect{\mu}_i\}$, i.e. the centers of RBFs, were fixed to a uniform grid in $\left[ 0, w_{max} \right]^2$, with $w_{max} = 65.0$ and $\sigma_i=12.0, ~\forall i$. Finally the weights $\gamma_i$ are related to the optimization variables $\rfp$ through a softmax transformation: 
\be
\gamma_i=\frac{e^{\varphi_i}}{\sum_{j=1}^Q e^{\varphi_j} }.
\ee
This is employed so that the resulting SDF integrates to $1$ which is  the variance of the corresponding Gaussian field.
We  made use of  a spectral representation of the underlying Gaussian field  (and therefore of $\bx_g$)  on the basis of its $\rfp$-controlled SDF and according to the formulations detailed in \cite{shinozuka1996simulation,hu1997simulation, sternfels2011stochastic}. The thresholded Gaussian vector $\bx_g$ gives rise to the binary microstructure  $\bx$ as described above and we denote summarily the corresponding transformation as:
\begin{align}
\bx=\bs{F}_{\rfp}(\bs{\Psi})
\label{eq:srx}
\end{align}
where $\bs{\Psi}$ denotes a vector of so-called random phase angles \cite{shinozuka1996simulation}. It consists of independent random variables uniformly distributed in $[0,2 \pi]$, and its dimension depends on  the discretization of the spectral domain.
A direct implication of Eq. \eqref{eq:srx}  is that the process-structure density $p(\bx|\rfp)$ can now be expressed as:
\be
p(\bx|\rfp) = \int \delta \left(\bx - \bs{F}_{\rfp}(\bs{\Psi})\right)~p(\bs{\Psi})~d\bs{\Psi}
\ee
where $p(\bs{\Psi})$ is the product of uniform densities $\mathcal{U}[0,2\pi]$.
As a result, expectations of arbitrary functions, say $f(\bx)$, with respect to $p(\bx|\rfp)$ can now be written as (with some abuse of notation):


\begin{align}
\mathbb{E}_{p \left( \bm{x} \middle| \varphi \right)} \left[ f \left( \bm{x} \right) \right] &= 
\int  f \left( \bm{x} \right) \delta \left( \bm{x} - \bm{F}_{\varphi} \left( \bm{\Psi} \right) \right) p \left( \bm{\Psi} \right) ~\diff \bs{x} ~\diff \bm{\Psi} \\
&= \int f \left( \bm{F}_{\varphi} \left( \bm{\Psi} \right) \right) p \left( \bm{\Psi} \right) \diff \bm{\Psi}
\label{eq:sup1}
\end{align}

\subsection*{VB-EM-Algorithm}

By making use of \refeqp{eq:sup1} above, we can write  the ELBO for the log-expected utility in \refeqp{eq:elbo1}  as

\begin{align}
\log U_{1, \mathcal{M}}^{\mathcal{D}}  \left( \varphi \right) &= \log \mathbb{E}_{p \left( \bm{x} \middle| \varphi \right)} \left[ u \left( \hp \right) p_{\mathcal{M}} \left( \hp \middle| \bm{x}, \mathcal{D} \right)  \diff \hp  \right] \nonumber \\
&= \log \int u \left( \hp \right) p_{\mathcal{M}} \left( \hp \middle| \bm{F}_{\varphi} \left( \bm{\Psi} \right), \mathcal{D}  \right) p \left( \bm{\Psi} \right) \diff \hp \diff \bm{\Psi} \nonumber \\
&\geq \mathbb{E}_{q_{\bm{\xi}} \left( \hp, \bm{\Psi} \right)} \left[ \log \frac{ u \left( \hp \right) p_{\mathcal{M}} \left( \hp \middle| \bm{F}_{\varphi} \left( \bm{\Psi} \right), \mathcal{D} \right) p \left( \bm{\Psi} \right)}{q_{\bm{\xi}} \left( \hp, \bm{\Psi} \right)} \right] \nonumber \\
&= \mathcal{F} \left( \varphi, q_{\bm{\xi}} \left( \hp, \bm{\Psi} \right) \right)
\end{align}

 where expectations with respect to $\bx$ have been substituted by integrations with respect to the (primal) random variables $\bs{\Psi}$ arising from the spectral representation. Similarly the variational density is expressed with respect to $\bs{\Psi}$, i.e. $q_{\bm{\xi}} \left( \hp, \bm{\Psi} \right)$ (as opposed to $ q_{\bm{\xi}} \left( \hp, \bm{x} \right)$). 
The ELBO of the (O2)-type problems in \refeqp{Eq:ELBO_O2} can similarly be written as
\begin{equation}
\begin{aligned}
    \log U_{2, \mathcal{M}}^{\mathcal{D}} \left( \varphi \right) &= \int p_{target} \left( \hp \right) \log p_{\mathcal{M}} \left( \hp \middle| \varphi, \mathcal{D} \right) \diff \hp \\
    &\approx \frac{1}{S} \sum\limits_{s=1}^S \log p_{\mathcal{M}} \left( \hp^{(s)} \middle| \varphi, \mathcal{D}  \right) \qquad \hp^{(s)} \stackrel{\text{\tiny{i.i.d.}}}{\sim} p_{target} \left( \hp \right)\\
    &= \frac{1}{S} \sum\limits_{s=1}^S \log \int p_{\mathcal{M}} \left( \hp^{(s)} \middle| \bm{F}_{\varphi} \left( \bm{\Psi} \right), \mathcal{D}  \right) p \left( \bm{\Psi} \right) \diff \bm{\Psi} \\
    &\geq  \frac{1}{S} \sum\limits_{s=1}^S \mathbb{E}_{q^{(s)}_{\bm{\xi}} \left( \bm{\Psi} \right)} \left[ \log \frac{  p_{\mathcal{M}} \left( \hp^{(s)} \middle| \bm{F}_{\varphi} \left( \bm{\Psi} \right), \mathcal{D}  \right) p \left( \bm{\Psi} \right) }{q^{(s)}_{\bm{\xi}} \left( \bm{\Psi} \right)} \right]  \\
    &= \sum\limits_{s=1}^S \mathcal{F}_s \left( q^{(s)}_{\bm{\xi}} \left( \bm{\Psi} \right), \varphi \right)
\end{aligned}
\end{equation}

Since the maximization of the ELBO w.r.t. $\bm{\xi}$ and $\varphi$ is not possible in closed form, noisy estimates of the gradients $\bm{\hat{g}}_{\varphi} \approx \nabla_\varphi \mathcal{F} \left( \varphi, q_{\bm{\xi}} \left( \hp, \bm{\Psi} \right) \right)$ and $\bm{\hat{g}}_{\bm{\xi}} \approx \nabla_{\bm{\xi}} \mathcal{F} \left( \varphi, q_{\bm{\xi}} \left( \hp, \bm{\Psi} \right) \right)$ are obtained
 using Monte Carlo and the reparametrization trick (\cite{kingma2013auto} -  see ensuing discussion).
 For our numerical illustrations,  the optimization with respect to  $\rfp$ and $\bm{\xi}$ is carried out with stochastic gradient ascent  and the Adam optimizer  \cite{kingma2014adam} in PyTorch \cite{paszke2019pytorch}.

\subsubsection*{Representation}

Instead of the bounded phase angles $\bs{\Psi}$, we employ the unbounded  and normally distributed variables $\bs{\Psi}_t$ (i.e. $\bs{\Psi}_t \sim \mathcal{N}( \bm{0}, \mat{I})$) which are related through the error function  $\text{erf} \left( \cdot \right)$ as follows: 

\begin{align}
    \Psi_i = 0.5 \cdot \left( 1 + \text{erf} \left( \frac{\Psi_{t,i}}{\sqrt{2 \pi}} \right) \right) \cdot 2 \pi 
\end{align}
As  a result, all expectations with respect to $p(\bs{\Psi})$ are substituted by expectations with respect to  $p \left( \bs{\Psi}_t \right) = \mathcal{N} \left( \bm{0}, \mat{I} \right)$.

In addition, instead of the Heaviside function defining the binary vector $\bx$ from the underlying Gaussian $\bx_g$ as $x_i=H(x_{g,i}-x_0)$, we employ the differentiable transformation 
\begin{align}
x_i=\frac{ \text{tanh} \left( \epsilon \left( x_{g,i}-x_0 \right) \right)+1}{2}
\end{align}  
We note that as  $\epsilon \to \infty$ we recover the Heaviside function and therefore a hard truncation. While the resulting $x_i$'s are only approximately binary for finite $\varepsilon$ (we used $\varepsilon=25$), these were used in all computations involved in the surrogate and the optimization. An advantage is that this enables the use of the reparametrization trick \cite{kingma2013auto} to estimate the ELBO and its gradients as explained in the previous section.

%
%

\subsubsection*{Low-rank Variational Approximation}

For $q_{\bm{\xi}} \left( \hp, \bs{\Psi}_t \right) \in \mathcal{Q}$  we adopt the choice of a low-rank multivariate Gaussian  distribution (with $\bm{z} = \left[ \hp, \bs{\Psi}_t \right]^T \in \mathbb{R}^{d_z}$), i.e.

\begin{align}
    q_{\vect{\xi}} \left( \vect{z} \right) = \mathcal{N} \left(\vect{z} ~| ~\vect{\mu}, \mat{\Sigma} = \text{diag} \left( \vect{d} \right) + \mat{L} \mat{L}^T  \right)
\end{align}

with $\mat{L} \in \mathbb{R}^{d_z \times M}$ and $M << d_z$. The variational parameters are given by $\bm{\xi} = \{ \bm{\mu}, \bm{d}, \mat{L} \}$ with $\text{dim} \left( \bm{\xi} \right) = \mathcal{O} \left( d_z \cdot M \right)$. This particular choice enables to capture enough of the correlation structure to drive the EM updates, while remaining scalable with regards to the (generally large) dimension $d_z$ of the problem (we used $M=50$).
We note that a fully Bayesian treatment of the surrogate could be accomplished by including the neural network parameters $\bm{\theta}$ in the variational inference framework.
While we could also drive the EM-algorithm via, e.g., Markov Chain Monte Carlo or Sequential Monte Carlo, the choice of variational inference is computationally faster and additionally enables monitoring of  convergence through the ELBO $\mathcal{F}$.











\subsubsection*{Tempering}

When specifying the material design objective, it is numerically advantageous to pursue a tempering schedule, in particular if the desired material behaviour deviates strongly from the initially observed dataset $\mathcal{D}$, or the   properties $\hp$  associated with the initial guess $\varphi^{(0)}$. In the following we  discuss an adaptive tempering strategy which - for the sake of illustration - we explain in the context of a utility function $u \left( \hp \right) = \mathbb{I}_{\mathcal{K}} \left( \hp \right)$ (see Fig. \ref{fig:objectivetypes-a}). Instead of trying to obtain $\varphi^* = \arg \max  p \left( \hp \in \mathcal{K} \middle| \varphi \right)$ directly, we instead introduce a sequence of target domains  $\mathcal{K}^{(r)}, r = 1..., R$, such that $\mathcal{K}^{(R)} = \mathcal{K}$. To this end we may define $\mathcal{K}^{(0)}$ in such a way, that (according to the model belief) a non-negligible number of samples $\hp \sim p_{\mathcal{M}} \left( \hp \middle| \varphi^{(0)}, \mathcal{D} \right)$ fall into the domain $\mathcal{K}^{(0)}$. In order to assess how strongly the tempered target domain $\mathcal{K}^{(r)}$ can be shifted towards the desired $\mathcal{K}$ in each each step $r$, we can make use of the \textit{effective sample size} (ESS) \cite{smith2013sequential}. For an ensemble of $N_w$ phase angles $\{ \bm{\Psi}^{(i)} \}_{i=1}^{N_w}$ generated from $q \left( \bm{\Psi} \right)$, we introduce the corresponding weights as the model-based belief that the material properties $\hp$ reside in the tempered target domain $\mathcal{K}^{(r)}$

\begin{align}
    w_n^r = \int \mathbb{I}_{\mathcal{K}^{(r)}} \left( \hp \right) p_{\mathcal{M}} \left( \hp \middle| \bm{F}_{\varphi} \left( \bs{\Psi}^{(i)}_t \right), \mathcal{D} \right) \diff \hp
\end{align}

Denoting the normalized weights as $\tilde{w}_n^r = w_n^r / \left( \sum_{n=1}^{N_w} w_n^r \right)$, the ESS is defined as 

\begin{align}
    \text{ESS} = \left( \sum_{n=1}^{N_w} \left. \tilde{w}_n^r \right. ^2 \right)^{-1}
 = \frac{\left( \sum_{n=1}^{N_w} w_n^r \right)^2}{\sum_{n=1}^{N_w} \left. w_n^r \right.^2}
\end{align}

where $\text{ESS} \in \left[ 0, 1 \right]$ represents the deterioration of sample quality induced by shifting the domain $\mathcal{K}^{(r)}$. Let  $q \left( \bs{\Psi}_t \right)$ be  an approximation to the posterior over the phase angles conditional on the optimality criteria (i.e. $\mathcal{K}^{(r)}$) and the current value of $\varphi$.  One may then adaptively chose to shift the target domain $\mathcal{K}^{(r)} \rightarrow \mathcal{K}^{(r+1)}$ in such a manner, that the ESS of the samples generated from $q \left( \bs{\Psi}_t \right)$ does not deteriorate beyond a certain threshold value (e.g. using a bisection approach). 
When defining the material design objective by means of a target distribution $p_{target} \left( \hp \right)$, similarly one may introduce tempering by gradually shifting the sample representation giving rise to the evidence lower bound.




\subsection*{Structure-Property linkage}

\subsubsection*{Probabilistic Surrogate}


The probabilistic surrogate employed 
 is based on a parametric convolutional neural network (see. Fig. \ref{fig:CNN_architecture}), where a split in the final dense layers gives rise to (separately) the mean vector $\bm{m}_{\bm{\theta}} \left( \bm{x} \right)$, as well as the covariance matrix $\mat{S}_{\bm{\theta}} \left( \bm{x} \right)$ (assumed to be diagonal). The specific choices made regarding the neural network architecture are based on prior published work (e.g. \cite{yang2018deep, cecen2018material}). Each block in Fig. \ref{fig:CNN_architecture} corresponds to 2d convolutions with a subsequent non-linear activation function (Leaky ReLU) ,followed by average pooling. The convolutional layers employ a $\left( 3 \times 3 \right)$ kernel, which in combination with appropriate padding leaves the size of the feature map unchanged \footnote{The discretized microstructures regarded as a vectors $\bm{x} \in \left\lbrace 0, 1 \right\rbrace^{4096}$ are of course reshaped into their original $\left( 64 \times 64 \right)$ un-flattened tensor representation for the CNN.}. The subsequent average pooling always employs a $ \left( 2 \times 2 \right)$ kernel (and identical stride), such that the size of the feature maps is reduced by half in each block. For the numerical results presented, after a sequence of $4$ such blocks (with an increasing depth of $4, 8, 12$ and $16$ channels in the feature maps), the resulting feature representation extracted from the microstructure is flattened and enters first a shared hidden layer (of width $30$), subsequently splitting up into two more layers that map to the mean $\bm{\mu}_{\bm{\theta}}$ and the diagonal covariance matrix $\mat{S}_{\bm{\theta}} \left( \bm{x} \right)$ (the positivity of the latter is ensured via an exponential transformation). The two phases were encoded as a $(+1)$ and $(-1)$ for the CNN, as this is numerically more expedient compared to an $\left\lbrace 1, 0 \right\rbrace$ representation of the phases. For all numerical experiments presented, the neural network was trained with a batch size of $N_{bs} = 128$. To add regularization, a weight decay of $10^{-5}$ was used, and additionally a dropout layer (with $p=0.05$) was introduced before the first dense layer. The neural network was trained on the log-likelihood of the data 
  making again use of the Adam optimizer for the stochastic updates of the parameters $\bm{\theta}$.

\subsubsection*{Physical model for the computation of properties $\hp$}

In the following we provide a more detailed description of the physical models in the  structure-property linkage abstractly represented   as $\hp ( \bm{x} )$, i.e., the link between the microstructures and their physical properties we want to control in this study. Note that the specific choice of  $\hp ( \bm{x} )$ does not have any direct bearing on the optimization, as the structure-property linkage only enters into the generation of the training data $\mathcal{D}$ for the surrogate. For our numerical illustrations we make use of both the effective \emph{thermal} as well as \emph{mechanical} properties of the microstructures $\bm{x} \sim  p \left( \bm{x} \middle| \varphi \right)$. We present details regarding the numerical computation of the effective mechanical properties, with the thermal properties following by analogy. In order to quantify the macroscopic response of a microstructure, we consider a linear, isotropic elasticity problem on the microscopic scale for a representative volume element (RVE). At this scale  the behaviour of the microstructure is characterized by the balance equation (BE) and constitute equation (CE)

\begin{align}
  &\text{(BE):} &   \text{div} \left( \mat{\sigma} \right) &= \vect{0} \qquad &&\forall \bs{s} \in  \Omega_{\text{RVE}}  \label{Eq:MicroscopicBE} \\
 &\text{(CE):}  &   \vect{\sigma} &= \mathbb{C} \left( \vect{s} \right) : \vect{\epsilon} &&\forall \bs{s} \in  \Omega_{\text{RVE}} \label{Eq:MicroscopicCE}
\end{align}

 where $\vect{\sigma}, \vect{\epsilon}$ denote microscopic stress and strain, while $\mathbb{C} \left( \vect{s} \right)$ constitutes the heterogeneous elasticity tensor $\mathbb{C} \left( \vect{s} \right)$. For a binary microstructures where the two  phases occupy (random) subdomains $\mathcal{V}_0$ and $\mathcal{V}_1$ (with $\mathcal{V}_0 \cap \mathcal{V}_1 = \emptyset$ and $\mathcal{V}_0 \cup \mathcal{V}_1 = \Omega_{\text{RVE}}$) the elasticity tensor follows as:

\begin{align}
    \mathbb{C} \left( \vect{s} \right) = 
    \begin{cases}
    \mathbb{C}_1, \qquad \qquad &\text{ if } \vect{s} \in \mathcal{V}_1 \\
    \mathbb{C}_0, & \text{ if } \vect{s} \in \mathcal{V}_0
    \end{cases}
\end{align}

In our case, the elasticity tensors  $\mathbb{C}_0$ and $\mathbb{C}_1$ are fully defined by the Young's moduli $E_0$ and $E_1$ of the two phases (a common Poisson's ratio of $\nu = 0.3$ was used). We define the \emph{macroscopic} stress $\mat{\Sigma} = \left\langle \vect{\sigma} \right\rangle$  as well as macroscopic strain $\mat{E} = \left\langle \vect{\epsilon} \right\rangle$, where $\left\langle \cdot \right\rangle$ denotes a \textit{spatial} average of microscopic quantities over $\Omega_{\text{RVE}}$. We then characterize the macroscopic, effective behaviour of the microstructure via \cite{miehe2002computational}

\begin{align}
    \mat{\Sigma} = \mathbb{C}^{\text{eff}} : \mat{E}
    \label{Eq:eff_behavior}
\end{align}

under the constraint that \eqref{Eq:eff_behavior} satisfies the averaging theorem by Hill \cite{miehe2002computational, hill1972constitutive} 

\begin{align}
    \mat{\Sigma} : \mat{E} = \frac{1}{\left| \Omega_{\text{RVE}} \right|} \int_{\partial \Omega_{\text{RVE}}} \vect{t} \cdot \vect{u} \diff A
\end{align}

with microscopic tractions $\bm{t}$ and displacements $\bm{u}$. The homogenized properties $\hp$ are based on various entries of $\mathbb{C}^{\text{eff}}$ which is computed by solving an ensemble of elementary load cases, as given by the following macroscopic strain modes

\begin{align}
    \mat{\hat{E}}^{(1)} =    \left[ {\begin{array}{cc}
   1 & 0 \\
   0 & 0 \\
  \end{array} } \right], ~~ 
  \mat{\hat{E}}^{(2)} =    \left[ {\begin{array}{cc}
   0 & 0 \\
   0 & 1 \\
  \end{array} } \right], ~~
  \mat{\hat{E}}^{(3)} =    \left[ {\begin{array}{cc}
   0 & 0.5 \\
   0.5 & 0 \\
  \end{array} } \right] 
\end{align}

corresponding to either a pure tension or shear mode, such that $\mathbb{C}^{\text{eff}}$ is recovered by integrating the microscopic stress $\bm{\sigma}$ to obtain $\mat{\Sigma} = \left\langle \bm{\sigma} \right\rangle$. One possible approach of imposing these elementary strain modes is based on the introduction of periodic boundary condition (as opposed to defining load cases based on displacement or tractions), which augments Eq. \eqref{Eq:MicroscopicBE} and \eqref{Eq:MicroscopicCE} by

\begin{align}
        \bm{\epsilon} &= \mat{\hat{E}}^{(c)} + \nabla_{\bm{s}} \bm{v} &&\text{ in } \Omega_{\text{RVE}} \label{Eq:AuxEq1}  \\
        \bm{v} &\phantom{=} && \text{ is $\Omega_{\text{RVE}}$-periodic} \label{Eq:AuxEq2} \\
       \bm{t} &= \bm{\sigma} \cdot \bm{n} && \text{ is $\Omega_{\text{RVE}}$-antiperiodic} \label{Eq:AuxEq3}
\end{align}

Here $\bm{v}$ denotes a periodic fluctuation (i.e., $\bm{u} = \bm{E} \bm{s} + \bm{v}$), and $\bm{t}$ are antiperiodic tractions on the boundary of the domain $\Omega_{\text{RVE}}$. We solve for the periodic fluctuations $\bm{v}$ for all three elementary load cases $c=\left\lbrace 1,2,3 \right\rbrace$ using the Bubnov-Galerkin approach and the standard Finite Element Method (an additional Lagrange multiplier has to be included in the variational problem to disambiguate it with regards to rigid body transformations). The effective tangent moduli $\mathbb{C}^{\text{eff}}$ of the RVE  thus obtained by the solution of the differential equations(Eq. \eqref{Eq:MicroscopicBE}, \eqref{Eq:MicroscopicCE}, \eqref{Eq:AuxEq1}, \eqref{Eq:AuxEq2} and \eqref{Eq:AuxEq3})  can be shown \cite{miehe2002computational} to satisfy the averaging theorem by Hill. We finally note  that $\mathbb{C}^{\text{eff}}$ and the  properties of interest  $\hp$  vary depending on the underlying, random microstructure $\bx$ and the need for their  repeated computation represents the computational bottleneck for the optimization of the process parameters $\varphi$.

\end{document}